%% file: main.tex
\documentclass[runningheads]{llncs}

\usepackage{eccv}

\usepackage{eccvabbrv}
\usepackage{graphicx}
\usepackage{booktabs}
\usepackage{multirow}
\usepackage{xcolor}
\usepackage[accsupp]{axessibility}
\usepackage{hyperref}
\usepackage{orcidlink}
\usepackage[normalem]{ulem}
\usepackage{soul}
\usepackage{placeins}
\usepackage{tikz}
\usetikzlibrary{positioning, fit, backgrounds, arrows.meta}

\input{defs}

\begin{document}

\title{STEP: Score-Based Temporal Energy for Human Pose Video Anomaly Detection}
\titlerunning{STEP: Score-Based Temporal Energy for Human Pose VAD}

\author{Jakub Micorek\inst{1}\orcidlink{0009-0000-5768-7980} \and
Mateusz Kozi\'nski\inst{2}\orcidlink{0000-0002-3187-518X} \and
Horst Possegger\inst{1}\orcidlink{0000-0002-5427-9938}}

\authorrunning{J.~Micorek et al.}

\institute{Institute of Visual Computing, Graz University of Technology, Austria \\
\email{\{jakub.micorek,possegger\}@tugraz.at}\\
\and
Institute for Medical Informatics, Statistics and Documentation, Medical University of Graz, Austria}

\maketitle

\begin{abstract}
\input{00_abstract}
\end{abstract}

\input{01_introduction}
\input{02_related_work}
\input{03_method}
\input{04_evaluation}
\input{05_discussion}
\input{06_conclusion}

\input{07_acknowledgements}

\clearpage
\bibliographystyle{splncs04}
\bibliography{main_refs}

\appendix
\clearpage

\newcommand{\beginsupplement}{
    \setcounter{section}{0}
    \renewcommand{\thesection}{\Alph{section}}
    \setcounter{figure}{0}
    \renewcommand{\thefigure}{S\arabic{figure}}
    \setcounter{table}{0}
    \renewcommand{\thetable}{S\arabic{table}}
    \setcounter{equation}{0}
    \renewcommand{\theequation}{S\arabic{equation}}
}
\beginsupplement

\begin{center}
    {\LARGE\bfseries Supplementary Material for:}\\[0.4em]
    {\large\bfseries STEP: Score-Based Temporal Energy for Human Pose Video Anomaly Detection}\\[0.2em]
\end{center}

\bigskip

\input{supplementary}

\end{document}

%% file: defs.tex
\definecolor{matcolor}{rgb}{1.0, 0.75, 0.79}

\newif\ifdraft
\draftfalse

\ifdraft

    \newcommand{\MK}[1]{\textcolor{matcolor}{[{\bf MK}: #1]}}
    
    \newcommand{\soutmk}[1]{\textcolor{matcolor}{\sout{#1}}}

    \newcommand{\JM}[1]{\textcolor{blue}{[{\bf JM}: #1]}}
    
    \newcommand{\soutjm}[1]{\textcolor{blue}{\sout{#1}}}

    \newcommand{\supp}[1]{\textcolor{magenta}{#1}}
    \newcommand{\soutsupp}[1]{\textcolor{magenta}{\sout{#1}}}

    \newcommand{\HP}[1]{\textcolor{green}{[{\bf HP}: #1]}}
    
    \newcommand{\southp}[1]{\textcolor{green}{\sout{#1}}}

\else
    
    \newcommand{\MK}[1]{}
    \newcommand{\soutmk}[1]{}

    \newcommand{\JM}[1]{}
    \newcommand{\soutjm}[1]{}

    \newcommand{\supp}[1]{}
    \newcommand{\soutsupp}[1]{}

    \newcommand{\HP}[1]{}
    \newcommand{\southp}[1]{}
\fi

\newcommand{\old}[1]{}

%% file: 00_abstract.tex
Skeleton-based Video Anomaly Detection (VAD) offers a robust, privacy-preserving solution for identifying abnormal behaviors. To model the distribution of normal static and moving poses, recent methods train Energy-Based Models (EBMs) via Denoising Score Matching (DSM). However, directly injecting noise, required for training, into raw joint coordinates creates physically impossible poses, and this structural collapse severely worsens as the temporal window expands.
To address this, we introduce STEP, a simple framework that utilizes Principal Component Analysis (PCA) to project pose sequences into a compact, whitened PC-space. Learning the data density within this well-behaved PC-space ensures that the injected noise translates into physically plausible variations, which allows the model to process longer video sequences without the performance collapse of raw coordinate baselines. Additionally, to mitigate inherent pose estimation inaccuracies arising from occlusions or motion blur, we integrate a sequence-level weighting mechanism based on the estimator's confidence scores.
Operating at real-time computational efficiency, our simple and lightweight framework outperforms the previous skeleton-based state-of-the-art by 12.2\% (90.1\% AUROC) on the challenging UBnormal dataset and achieves highly competitive results by improving on the ShanghaiTech benchmark.

%% file: 01_introduction.tex
\section{Introduction}
\label{sec:intro}

\begin{figure}[tb]
  \centering
  \includegraphics[width=\textwidth]{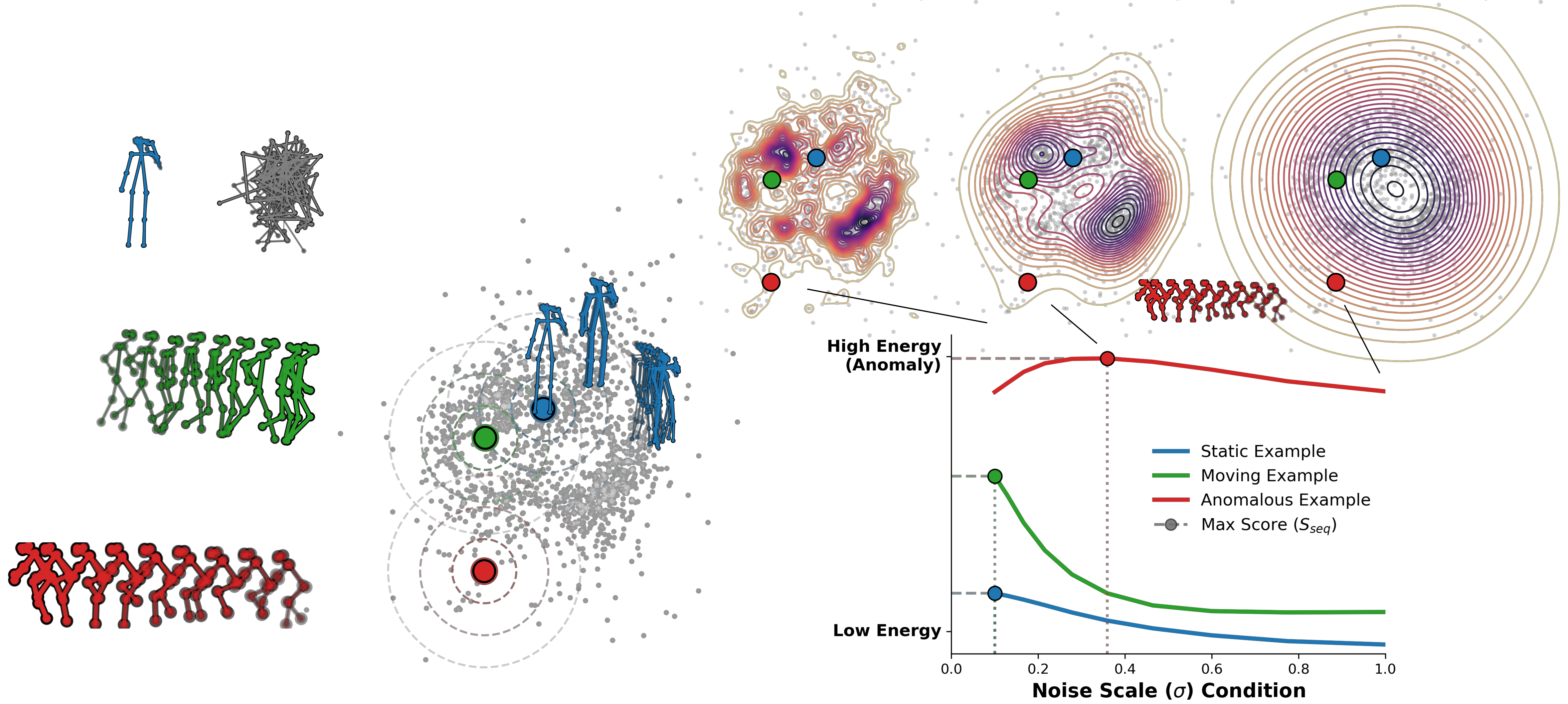}
  \caption{\textbf{The STEP Pipeline.} \textbf{Left:} Human poses are extracted from the raw video. We highlight a static person (blue), a walking person (green), and an anomalous cyclist (red). Naively adding independent Gaussian noise to pixel coordinates on the static person destroys the kinematic structure, creating physically impossible poses (gray). \textbf{Middle:} In contrast, by projecting the sequences into a compact, whitened PC-space, the injected isotropic noise translates into semantically meaningful action variations (\eg, smoothly inducing motion in the static blue pose sequence) while yielding structurally plausible bone lengths and human proportions. \textbf{Right:} Consequently, the Energy-Based Model learns smoothed, well-behaved energy landscapes at different $\sigma$ scales (top). During inference, we evaluate the sequence across distinct noise scales ($\sigma_{low}$ to $\sigma_{high}$). Normal actions (static and walking) fall into lower-energy basins across these scales, while the anomalous cyclist yields a high-energy score.}
  \label{fig:pipeline}
\end{figure}

Detecting anomalies in human motion sequences is pivotal for enabling a timely response to emergencies: accidents in production facilities, patient falls in elderly care institutions, or acts of violence in public spaces. However, algorithmic detection of anomalous motion sequences remains an open problem. Its inherent challenge stems from the lack of anomalous training data: only normal motion sequences are available during training. This prohibits framing the problem as event detection or motion sequence classification.

\looseness=-1
The classical anomaly detection approach relies on training a deep model in self-supervised tasks, like auto-encoding motion sequences~\cite{georgescu2021anomaly, georgescu2021background, hasan2016learning, ionescu2019object} or predicting masked data fragments \cite{liu2018future, nguyen2019anomaly,georgescu2021anomaly}. Training data is anomaly-free, but at test time, the model performs the task on possibly anomalous data items, and a task-specific error measure is used as an anomaly indicator. Unfortunately, this method is limited by an inherent flaw: it is predicated on the assumption that a model trained on normal data fails for anomalous input. In reality, deep networks generalize beyond their training set, and there is no guarantee that all anomalies deteriorate the network's performance.

Recently, a less heuristic approach has received increasing attention: both training and non-anomalous test data are treated as instantiations of a random variable with a fixed probability distribution, and anomaly detection is posed as estimating the energy of each test item~\cite{micorek24cvpr}. To that end, an energy-based model is trained on anomaly-free training data and used to predict the energy of the possibly anomalous test items. Test samples with high estimated energy, in other words, ones unlikely under the model of the non-anomalous data distribution, are deemed anomalous. Samples with low estimated energy are considered normal. The advantage of this approach is that it casts the practical challenge of accurate anomaly detection into a mathematically well-defined objective of developing accurate approximations to the energy function.

\looseness=-1
Unfortunately, training energy-based anomaly detectors remains challenging. A typical approach is to harness the Denoising Score Matching (DSM)~\cite{vincent2011connection} objective, which relies on injecting noise into training samples. However, naively applying DSM to human pose sequences, with their complex pattern of joint dependencies, displays two fundamental deficiencies. First, as illustrated in the left part of \cref{fig:pipeline}, injecting isotropic Gaussian noise into joint coordinates destroys the kinematic structure of the human body: bone lengths warp, symmetries break, and the sequence degrades into a cloud of independent joints. Crucially, this structural collapse compounds as the temporal window $T$ expands. This drastically inflates the space of physically impossible poses, thereby severely degrading the model's ability to learn a meaningful energy landscape. Consequently, the expressive power of the energy model is wasted on modeling impossible body configurations. Second, existing approaches treat pose sequences extracted from images as perfect ground truth. In reality, off-the-shelf pose estimators applied to real-world surveillance footage suffer from coordinate jitter, missing joints, and occlusion, which causes current models to warp energy landscapes around tracking failures.

To address these deficiencies, we propose Score-based Temporal Energy for Poses (STEP), a novel energy-based approach to anomaly detection in human motion sequences. STEP is trained with Denoising Score Matching (DSM). However, in contrast to previous approaches leveraging DSM, STEP foregoes modeling the data distribution in the original space of joint coordinate sequences in favor of a compact space resulting from Principal Component Analysis (PCA). As shown in the middle of \cref{fig:pipeline}, injecting isotropic noise into this representation perturbs human motion sequences along learned basis trajectories, generating semantically meaningful action variations while preserving natural human proportions. As a result, the expressive power of the distribution model is spent on representing feasible motion sequences, as opposed to kinematically impossible noise. Consequently, as depicted in the right part of \cref{fig:pipeline}, our Energy-Based Model learns a smoothed, well-behaved energy landscape where normal actions naturally fall into low-energy basins.

Furthermore, to account for the inherent unreliability of pose estimators, we introduce a soft confidence weighting mechanism into the DSM training objective and test-time energy estimation. This discounts the impact of tracking failures without perturbing the temporal data structure or provoking the omission of critical anomalies, yielding a robust and accurate anomaly detector.

The technical contributions behind STEP are summarized as follows:
\begin{itemize}
    \item \textbf{PC-Space Density Estimation:}
    We demonstrate that naively applying Denoising Score Matching to raw coordinates yields suboptimal performance and worsens as the temporal dimension $T$ grows. We resolve this by operating in a compact, whitened PC-space, ensuring noise injection translates to meaningful pose sequence variations.
    \item \textbf{Confidence-Aware Anomaly Scoring:} We introduce a sequence-level confidence weighting mechanism into the DSM framework, both at training and test time. This accounts for low-confidence poses and provides a significant boost to overall AUROC.
    \item \textbf{Architectural Conditioning:} We show that introducing a $\sigma$-modulated Residual MLP, which utilizes skip-connections and per-layer noise conditioning, consistently outperforms the architectures used in previous work.
\end{itemize}

In exhaustive experiments on two challenging human motion datasets, UBnormal and ShanghaiTech, the STEP framework outperforms the previous skeleton-based state-of-the-art by 12.2\% and 0.3\% AUROC, respectively. Crucially, these performance gains require negligible computational overhead. Our lightweight architecture processes dense crowds in under 1\,ms, leaving overall pipeline latency dominated solely by the upstream pose extraction.

%% file: 02_related_work.tex
\section{Related Work}
\label{sec:related_work}

\looseness=-1
Anomaly detection in human motion sequences has become a prominent focus, as it abstracts away the background clutter and illumination changes that typically challenge traditional pixel-space approaches. Historically, one way to detect anomalies in human motion is directly from raw videos. This task, often framed as Video Anomaly Detection (VAD) under one-class classification (OCC), relies on training exclusively on normal data~\cite{ucfcrime_sultani_2018_real_world_vad, zaheer2022generative, ubnormal_Acsintoae_CVPR_2022, Yan_2023_ICCV}. These standard approaches typically categorize into frame-centric models analyzing global features~\cite{wang2019gods, Yan_2023_ICCV, zaheer2022generative} and object-centric models estimating bounding box abnormalities~\cite{ionescu2019object, georgescu2021anomaly, wang2022video, BARBALAU2023103656, Flaborea_2023_ICCV}. Such methods often solve proxy tasks like auto-encoding or future-frame prediction~\cite{hasan2016learning, shanghaitech_luo2017revisit, park2020learning, liu2018future, yu2020cloze}. However, multimodal and appearance-based modeling from videos remains vulnerable to demographic bias, privacy concerns, and scene-specific artifacts, motivating a strong shift toward purely skeletal kinematics.

Early skeleton-based methods framed motion anomaly detection primarily as an unsupervised reconstruction or prediction problem~\cite{morais2019cvpr, li2021neurocmp, jain2020icpr, rodrigues2020wacv, markovitz2020cvpr, miracle2022arxiv}. However, standard deep networks tend to over-generalize, successfully reconstructing even unseen anomalous behaviors. To better capture the multimodal diversity of human motion, contemporary models have increasingly adopted explicit probabilistic density estimation. For instance, while MoCoDAD~\cite{Flaborea_2023_ICCV} utilizes diffusion models to forecast future poses, it ultimately falls back on measuring coordinate-level reconstruction errors to score anomalies rather than evaluating the true probability of the sequence. Conversely, frameworks that do compute true probabilities rely on highly specialized architectures to model skeletal dependencies before applying density estimation. For instance, alongside earlier spatial networks like Normal Graph~\cite{luo2020neurcomp} and COSKAD~\cite{flaborea2024pr}, STG-NF~\cite{Hirschorn_2023_ICCV} couples Spatio-Temporal Graph Convolutional Networks (ST-GCNs) with Normalizing Flows, whereas SeeKer~\cite{delic2025seeker} pairs autoregressive keypoint factorization with explicit per-keypoint likelihoods.
Alternatively, EBMs trained via DSM provide a highly adaptable approach to density estimation~\cite{hyvarinen05, vincent2011connection, song2019generative}. As demonstrated by MULDE~\cite{micorek24cvpr}, this feature-agnostic technique can be applied directly to multimodal video features and individual poses. However, as discussed, naively applying this framework to raw coordinates is fundamentally flawed due to noise-induced structural collapse. Instead, STEP ensures robust density estimation by operating in a structure-preserving, whitened PC space.

%% file: 03_method.tex
\section{Method}
\label{sec:method}

\paragraph{Problem formulation}
We are given a training set $D$ of human motion sequences extracted by an off-the-shelf pose tracker from a video recording.
Each sequence $X\in D$ takes the form of a tensor $X\in \mathbb{R}^{2 \times V \times T}$, where the first dimension corresponds to the two image coordinates, $V$ is the number of joints, and $T$ is the number of frames included in the sequence.
Our goal is to train an anomaly detector to classify test sequences of the same form as either normal or anomalous.

\paragraph{Overview of our approach}
We assume that both training samples and non-anomalous test samples originate from the same data distribution
and train a deep network $f$ to estimate sample energy.
At test time, we use $f$ as an anomaly indicator.
Our training procedure relies on denoising score matching (DSM), which we introduce in \cref{sec:dsm}.
However, since a direct application of DSM to human motion sequences results in subpar performance,
we propose modifications:
As described in \cref{sec:pca}, we model the distribution of the motion sequences in the space of principal components, as opposed to the space of raw joint coordinates;
We introduce pose confidence weighting into training and prediction, as outlined in \cref{sec:confidence},
and verify a new architecture of the energy-based model, described in \cref{sec:architecture}.

\subsection{Denoising score matching for anomaly detection}
\label{sec:dsm}
Our work builds on denoising score matching for anomaly detection~\cite{micorek24cvpr}.
Denoising score matching~\cite{vincent2011connection} is a formulation for learning the energy gradient of a distribution given in the form of a data sample,
and~\cite{micorek24cvpr} harnessed this approach for learning the energy function itself.
Below, we first introduce DSM and then describe its use for anomaly detection in previous work.

Given a training set $D$ of data items $\mathbf{x}$ drawn from a fixed distribution with probability density $p$,
DSM lets one train a deep network to approximate the energy gradient of the training distribution injected with noise.
Formally, the noise-injected distribution is defined as
\begin{equation} \label{eq:noisy_distribution}
q (\tilde{\mathbf{x}}) = \int \rho(\tilde{\mathbf{x}} \vert \mathbf{x} ) p(\mathbf{x}) d \mathbf{x} ,
\end{equation}
where $\rho(\tilde{\mathbf{x}} \vert \mathbf{x} )$ denotes the conditional distribution of a noisy sample $\tilde{\mathbf{x}}$ given a noise-free sample $\mathbf{x}$, universally taken to be an \iid Gaussian centered at $\mathbf{x}$.
The main idea behind DSM is that $q$ preserves the shape of $p$ but leads to a simpler and more effective training formulation.
The energy of a sample $\tilde{\mathbf{x}}$ is defined as $E_q (\tilde{\mathbf{x}}) = - \log q(\mathbf{\tilde{x}})$.
DSM trains a neural network $s_\theta$, parameterized with a vector $\theta$, to approximate the energy gradient $\nabla_{\mathbf{\tilde{x}}} E_q (\tilde{\mathbf{x}}) $ in the sense of solving
\begin{equation} \label{eq:noised_score_matching}
\min_\theta \mathbb{E}_{\mathbf{\tilde{x}}\sim q(\mathbf{\tilde{x}})} \left\| s_{\theta}(\mathbf{\tilde{x}}) - \nabla_{\mathbf{\tilde{x}}} E_q (\tilde{\mathbf{x}}) \right\|_2^2 .
\end{equation}
Directly evaluating~\eqref{eq:noised_score_matching} is impossible because $q(\mathbf{\tilde{x}})$ is not known analytically, but Vincent et al.~\cite{vincent2011connection} showed that it is equivalent, up to an additive constant, to
\begin{equation} \label{eq:score_matching_gaussian}
\min_\theta
\mathbb{E}_{\substack{\mathbf{x}            \sim p(\mathbf{x}) \\
                      \mathbf{\tilde{x}}    \sim \mathcal{N}(\mathbf{\tilde{x}} | \mathbf{x},\sigma \mathbf{I})}}
\left\| s_{\theta}(\mathbf{\tilde{x}}) -
  \frac{\mathbf{\tilde{x}} - \mathbf{x}}{\sigma^{2}}
  \right\|_2^2 ,
\end{equation}
which can be evaluated efficiently.
DSM owes its name to the fact that training $s$ with the objective~\eqref{eq:score_matching_gaussian} resembles training the model for denoising.

Detecting anomalies requires approximating the energy $E_q(\tilde{\mathbf{x}})$, as opposed to its gradient.
To that end, recent work~\cite{micorek24cvpr} modified the DSM objective~\eqref{eq:score_matching_gaussian} to train the gradient of $f$, instead of the network itself,
\begin{equation}
\label{eq:training_gradient}
\min_\theta
\mathbb{E}_{\hspace{-3mm}\substack{\mathbf{x}            \sim p(\mathbf{x}) \\
                      \mathbf{\tilde{x}}    \sim \mathcal{N}(\tilde{\mathbf{x}} | \mathbf{x},\sigma \mathbf{I})}}
   \left\|
      \nabla_{\mathbf{\tilde{x}}} f_\theta \left(\mathbf{\tilde{x}}\right)
      -
      \frac{\mathbf{\tilde{x}} - \mathbf{x}}{\sigma^2}
   \right\|_2^2
,
\end{equation}
making $\nabla_{\mathbf{\tilde{x}}} f_\theta \left(\mathbf{\tilde{x}}\right)$ approximate $\nabla_{\tilde{\mathbf{x}}} E_q(\tilde{\mathbf{x}})$,
thereby aligning $f$ with $E_q(\tilde{\mathbf{x}})$ up to a constant bias.
In practice, instead of approximating $E_q(\tilde{\mathbf{x}})$ for a fixed noise scale $\sigma$, the network $f_\theta$ is trained to approximate a family of energies $E_{q_\sigma} (\tilde{\mathbf{x}})$, parameterized by $\sigma$, with a multi-scale objective
\begin{equation}
    \min_{\theta} \mathbb{E}_{\substack{\mathbf{x} \sim p(\mathbf{x}) \\ \mathbf{\tilde{x}} \sim \mathcal{N}( \mathbf{\tilde{x}} | \mathbf{x}, \sigma^2 I) \\ \sigma \sim \mathcal{U}(\{\sigma_i\}_{i=1}^L)}}
         \lambda(\sigma) \left\| \nabla_{\mathbf{\tilde{x}}} f_\theta(\mathbf{\tilde{x}}, \sigma) - \frac{\mathbf{\tilde{x}} - \mathbf{x}}{\sigma^2} \right\|_2^2 \;,
\end{equation}
where $\mathcal{U}(\{\sigma_i\}_{i=1}^L)$ denotes a uniform distribution over $L$ predefined noise scales $\sigma_i$, and $\lambda(\sigma)=\sigma^2$ is a scale-specific weight.
At test time, energy estimates corresponding to different noise levels are aggregated into a single anomaly indicator.

\subsection{Denoising score matching in the space of principal components}
\label{sec:pca}
The disadvantage of DSM in the context of human motion sequences is that injecting \iid Gaussian noise into sequences of joint coordinates ignores the structure of the human body
and destroys the pattern of statistical dependencies between joint positions.
The resulting probability distribution assigns mass to kinematically implausible motion patterns,
like the one shown in gray in the left part of \cref{fig:pipeline}.
This defeats the purpose of training the model to capture the manifold of human motion:
Modelling capacity is spent on accommodating implausible poses instead of outlining the limits of normal, kinematically feasible sequences.

We address the problem by moving away from the original space of joint coordinates and modeling the distribution of the motion sequences in a more compact space, where Gaussian noise corresponds to valid poses, as opposed to kinematically implausible ones. We tested three alternative techniques of projecting motion sequences to such a space: Principal Component Analysis, a plain autoencoder and a variational autoencoder.
We selected PCA, because projecting motion sequences to the whitened space of principal components (PC) yielded the highest anomaly detection accuracy (see detailed experiments in the supplementary material).
To formalize this approach, we
denote the projection matrix of the top $K$ principal components, estimated on the training set, by $W_K$, the diagonal matrix of the corresponding eigenvalues by $\Lambda_K$ and the training data mean by $\mu$,
and formalize the projection to the whitened PC space as:
\begin{equation}
    \pi(\mathbf{x}) = \Lambda_K^{-\frac{1}{2}} W_K^T (\mathbf{x} - \mu) \;.
    \label{eq:whitening}
\end{equation}
The projection effectively scales the latent space into an isotropic hypersphere.

\input{pc_interpretation}

Modelling the distribution of motion sequences in the whitened space of principal components has two advantages.
First, injecting noise into this representation results in plausible human motion sequences, like the one shown in blue in the left part of \cref{fig:pipeline}.
In \cref{fig:pca_traversal}, we show that principal components represent meaningful actions, like walking.
Second, PCA acts as a denoiser: limiting the representation to top $K$ PCs removes high-frequency motion,
letting the energy model focus on the overarching motion trajectory instead of frame-to-frame fluctuation of joint positions.
Our experiments in \cref{sec:ablation} show that moving the model to the whitened space results in a considerable performance boost.

\subsection{Accounting for uncertainty in pose extraction}
\label{sec:confidence}
The DSM training objective~\eqref{eq:training_gradient} treats all data items equally.
In our context of human motion anomaly detection, this ignores the fact that pose sequences extracted from videos are associated with uncertainty.
In consequence, the capacity of the energy model is spent on trying to accommodate erroneous poses.

To ensure the learned energy landscape is more robust against severe tracking and detection failures, we discount the influence of uncertain pose sequences through confidence weighting.
We denote an average joint position confidence score provided by the pose estimator for motion sequence $\mathbf{x}$ by $c(\mathbf{x}) \in [0, 1]$.
We use $c(\mathbf{x})$ to reweight loss terms corresponding to individual training sequences.
To that end, we define a weighted multi-scale DSM loss as
\begin{equation}
    \min_{\theta} \mathbb{E}_{\substack{\mathbf{x} \sim p(\mathbf{x}) \\ \mathbf{z} \sim \mathcal{N}( \mathbf{z} | \pi(\mathbf{x}), \sigma^2 I) \\ \sigma \sim \mathcal{U}(\{\sigma_i\}_{i=1}^L)}}
        c(\mathbf{x}) \lambda(\sigma) \left\| \nabla_{\mathbf{z}} f_\theta(\mathbf{z}, \sigma) - \frac{\mathbf{z} - \pi(\mathbf{x})}{\sigma^2} \right\|_2^2 \;,
    \label{eq:loss_confidence}
\end{equation}
where $\mathbf{z}$ denotes a noisy vector in the PC space.
$c(\mathbf{x})$ downweights low-confidence measurements, preventing tracking failures from distorting the energy landscape.

At test-time, given a motion sequence $\mathbf{x}$ and a sequence of pre-defined noise levels $\sigma_i$, we use the trained energy model $f$ to compute noise-level-specific energy estimates $f(\pi(\mathbf{x}),\sigma_i)$.
To aggregate them into a single anomaly estimator, we standardize the energy at the $i$-th noise level using the mean energy $\bar{E}_i$ and energy variance $\mathrm{Var}(E_i)$, computed over the training set, and then select the largest standardized energy estimate.
To accommodate the uncertainty of the pose extractor also at test time, we weight the standardized energy estimates by the confidence estimate $c$. Formally, the aggregated anomaly indicator is
\begin{equation}
\label{eq:aggregation}
A(\mathbf{x}) = c(\mathbf{x}) \max_{i} \frac{f(\pi(\mathbf{x}), \sigma_i) - \bar{E}_i}{\sqrt{\mathrm{Var}(E_i)}}.
\end{equation}
Our ablation study shows that discounting the anomaly score of uncertain motion sequences often prevents a common error consisting of declaring a sequence anomalous due to tracking errors.

\subsection{Architecture of the Energy-Based Model}
\label{sec:architecture}
Previous skeleton-based energy models, such as MULDE~\cite{micorek24cvpr}, typically rely on standard MLPs where the noise scale $\sigma$ is only integrated via early-fusion concatenation at the input. We argue that this leaves the architectural design space for noise-dependent density estimation largely unexplored. Instead, we propose a $\sigma$-modulated Residual MLP that treats $\sigma$ as a global conditioning signal. As shown in \cref{fig:architecture}, $\sigma$ is explicitly routed to every hidden layer to modulate activations within each residual block. Our ablation study (\Cref{tab:unified_ablation}) confirms that this dense conditioning approach consistently outperforms the vanilla MLP baseline.

\begin{figure}[b]
\centering
\resizebox{\textwidth}{!}{
\begin{tikzpicture}[
    >=stealth,
    base/.style={draw, align=center, minimum height=0.7cm, rounded corners, thick, font=\small},
    data/.style={base, fill=blue!10, draw=blue!80!black},
    sigma/.style={base, fill=orange!10, draw=orange!80!black},
    concat/.style={base, draw=gray!80!black},
    block/.style={base, fill=gray!10, draw=gray!80!black, minimum width=1.4cm},
    math/.style={circle, draw, thick, inner sep=1pt, minimum size=0.4cm, fill=white},
    data_arrow/.style={->, thick, blue!80!black},
    sigma_arrow/.style={->, thick, orange!80!black},
    grad_arrow/.style={->, thick, densely dashdotted, green!45!black},
    act/.style={base, fill=white, draw=black, minimum height=0.4cm, font=\scriptsize\bfseries\color{black}},
]

\node (z) [align=center] {{$\pi(\textbf{x})$}};
\node (concat) [concat, right=0.8cm of z, label={[label distance=0.08cm] below:\footnotesize Early Fusion}] {$[\pi(\mathbf{x}), \sigma]$};
\node (lin1) [data, right=0.5cm of concat] {Linear};
\node (b1) [block, right=0.5cm of lin1] {Block 1};
\node (dots) [right=0.3cm of b1, font=\Large] {$\cdots$};
\node (bn) [block, right=0.3cm of dots] {Block $N$};
\node (final) [data, right=0.5cm of bn] {Linear};
\node (out) [right=0.6cm of final, align=center] {\textbf{$f_\theta(\pi(\mathbf{x}), \sigma)$}};

\node (sigma_label) [above=1cm of z, font=\bfseries] {$\sigma$};
\coordinate (sigma_dash_start) at (b1.east |- sigma_label);
\coordinate (sigma_dash_end)   at (bn.west |- sigma_label);
\coordinate (sigmaline_end)    at (bn.north |- sigma_label);

\draw [data_arrow] (z) -- (concat);
\draw [data_arrow] (concat) -- (lin1);
\draw [data_arrow] (lin1) -- (b1);
\draw [data_arrow] (b1) -- (dots);
\draw [data_arrow] (dots) -- (bn);
\draw [data_arrow] (bn) -- (final);
\draw [data_arrow] (final) -- (out);

\draw [thick, orange!80!black] (sigma_label) -- (sigma_dash_start); 
\draw [thick, orange!80!black, dashed] (sigma_dash_start) -- (sigma_dash_end);
\draw [thick, orange!80!black] (sigma_dash_end) -- (sigmaline_end);
\draw [sigma_arrow] (sigma_label -| concat) -- (concat.north);
\draw [sigma_arrow] (sigma_label -| b1) -- (b1.north);
\draw [sigma_arrow] (sigmaline_end) -- (bn.north);

\coordinate (grad_start) at ([yshift=-1.2cm]out.south);
\draw [grad_arrow] (out.south) -- (grad_start) -| (z.south) node[pos=0.25, yshift=-0.04cm, below, font=\footnotesize\bfseries] {$\nabla_{\pi(\mathbf{x})} f_\theta(\pi(\mathbf{x}), \sigma)$};

\node (lin_micro) [data, right=2cm of out, yshift=0.9cm] {Linear};
\node (enc) [sigma, minimum height=0.4cm, right=0.7cm of lin_micro, yshift=0.6cm] {Linear};
\node (alpha) [sigma, minimum height=0.4cm, right=0.4cm of enc, yshift=-0.8cm] {$\alpha(\sigma)$};

\path (enc) -- (alpha) coordinate[midway] (sigma_mid);

\node (gelu) [act, below=0.3cm of lin_micro] {GELU};
\node (add1) [math, below=0.7cm of gelu] {$\oplus$};
\node (gelu_sigma) [act, below=0.2cm of enc] {GELU};
\node (lin2_sigma) [sigma, minimum height=0.4cm, below=0.2cm of gelu_sigma] {Linear};
\node (ln_sigma) [act, below=0.2cm of lin2_sigma] {LayerNorm};
\node (mult) [math] at (enc |- add1) {$\otimes$};
\node (add2) [math, below=0.2cm of add1] {$\oplus$};

\coordinate (fin_start) at ([yshift=1.3cm]lin_micro.north);
\coordinate (sigma_center_top) at ([yshift=0.9cm]sigma_mid);
\coordinate (sigma_start) at (fin_start -| sigma_center_top);

\coordinate (box_top) at ([yshift=1.1cm]lin_micro.north);
\coordinate (box_bottom) at ([yshift=-0.3cm]add2.south);
\coordinate (box_left)  at ([xshift=-1.45cm]lin_micro.west); 
\coordinate (box_right) at ([xshift=0.25cm]alpha.east);

\draw [data_arrow] (fin_start) -- (lin_micro);
\draw [sigma_arrow] (sigma_start) -- (sigma_center_top) node[pos=0.6, right, font=\bfseries] {$\sigma$};

\draw [sigma_arrow] (sigma_center_top) -| (enc);
\draw [sigma_arrow] (sigma_center_top) -| (alpha);
\draw [sigma_arrow] (enc) -- (gelu_sigma);
\draw [sigma_arrow] (gelu_sigma) -- (lin2_sigma);
\draw [sigma_arrow] (lin2_sigma) -- (ln_sigma);
\draw [sigma_arrow] (ln_sigma) -- (mult);
\draw [sigma_arrow] (alpha) |- (mult) node[pos=0.7, below, font=\scriptsize] {Scale by $\alpha$};

\draw [data_arrow] (lin_micro) -- (gelu);
\draw [data_arrow] (gelu) -- (add1);
\draw [sigma_arrow] (mult) -- (add1) node[midway, below=0.05cm, font=\scriptsize\color{orange!80!black}] {Modulation};
\draw [data_arrow] (add1) -- (add2);

\coordinate (fout_end) at ([yshift=-0.6cm]add2.south);
\draw [data_arrow] (add2) -- (fout_end);

\coordinate (res_branch) at ([yshift=0.5cm]lin_micro.north);
\fill [blue!80!black] (res_branch) circle (1.5pt);
\coordinate (res_mid) at ([xshift=-1.0cm]lin_micro.west);
\draw [data_arrow] (res_branch) -| (res_mid) |- (add2.west) node[near end, below=0.1cm, font=\scriptsize, color=blue!80!black] {Residual Path};

\begin{scope}[on background layer]
    \node [draw=gray!80, thick, fill=gray!5, rounded corners, fit=(box_top) (box_bottom) (box_left) (box_right), inner sep=0cm] (microbox) {};
\end{scope}

\node [anchor=south east, font=\bfseries\color{gray!80!black}, inner sep=0.15cm] at (microbox.south east) {Block};

\draw [dashed, thick, gray!30] (bn.north east) -- (microbox.north west);
\draw [dashed, thick, gray!30] (bn.south east) -- (microbox.south west);

\end{tikzpicture}
}
\caption{\textbf{STEP Architecture.} \textbf{Left:} Overview of our energy model. The noise scale $\sigma$ is integrated via early fusion with the whitened pose sequence $\pi(\mathbf{x})$ and additionally input to every stacked block. The network $f_\theta(\pi(\mathbf{x}), \sigma)$ outputs a scalar energy estimate. The dashed green path illustrates the energy gradient calculation, which is required only during training. \textbf{Right:} Internal structure of the modulated block, combining conditioning on $\sigma$ with a residual skip connection.}
\label{fig:architecture}
\end{figure}

%% file: pc_interpretation.tex
\begin{figure}[tb]
  \centering
  \includegraphics[width=\textwidth]{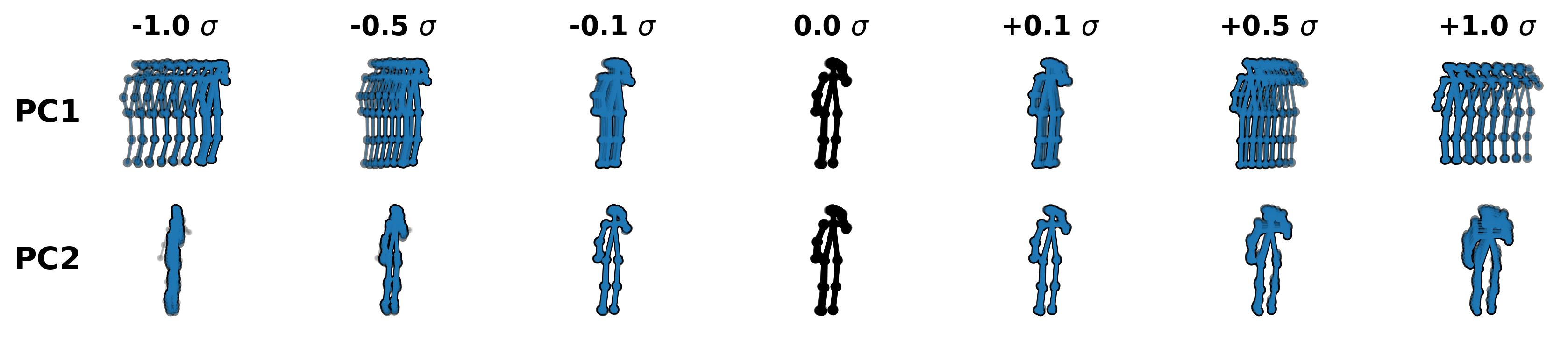}
  \caption{
\textbf{The principal components correspond to semantically meaningful actions}:
The image shows the effect of shifting a static pose along the first two principal components.
PC1 corresponds to walking across the field of view, while PC2 represents walking towards or away from the camera.
}
  \label{fig:pca_traversal}
\end{figure}

%% file: 04_evaluation.tex
\section{Experiments}
\label{sec:experiments}
To validate the effectiveness of our approach, we evaluate our STEP framework, comparing its performance against the state-of-the-art, ablating its core architectural components, and analyzing its real-time computational efficiency.

\subsection{Experimental Setup}

\textbf{Datasets.} We evaluate our method on two popular and challenging Video Anomaly Detection benchmarks: ShanghaiTech~\cite{shanghaitech_luo2017revisit} and UBnormal~\cite{ubnormal_Acsintoae_CVPR_2022}. ShanghaiTech is a real-world dataset comprising 330 training and 107 test videos across 13 scenes, featuring anomalous events such as robbery, fighting, and cycling. UBnormal is a synthetic, photorealistic dataset composed of 543 videos (268 training, 64 validation, and 211 test sequences) spanning 29 virtual scenes. It introduces complex human-related anomalies, such as fighting and running, alongside non-human and environmental anomalies, and is highly accurately labeled. Because both datasets contain events that are not explicitly driven by human pose (\eg, fire or weather), we follow established protocols~\cite{Flaborea_2023_ICCV, rodrigues2020wacv} and report results on both the Full test and the Human-Related (HR) splits, which isolate anomalies directly involving human actions.

\textbf{Baselines.} We compare STEP against a broad range of skeleton-based VAD algorithms. While we compare our framework to a wide range of traditional reconstruction and prediction models~\cite{morais2019cvpr, markovitz2020cvpr, miracle2022arxiv, rodrigues2020wacv, jain2020icpr, luo2020neurcomp, flaborea2024pr, li2021neurocmp, Flaborea_2023_ICCV}, our primary focus is on probabilistic density estimation methods. While STG-NF~\cite{Hirschorn_2023_ICCV} and SeeKer~\cite{delic2025seeker} define the current performance ceiling on ShanghaiTech and UBnormal for skeleton-based VAD, MULDE~\cite{micorek24cvpr} serves as our most direct methodological baseline. By exposing the structural collapse inherent in MULDE's DSM framework when applied to raw coordinates, we demonstrate how our PC-space and confidence weighting approach effectively resolves these fundamental limitations. A broader comparison against additional multi-modal and pixel-based methods is provided in the supplementary material.

\textbf{Evaluation Metrics.} We use the standard Area Under the Receiver Operating Characteristic curve (AUROC) as our evaluation metric. To ensure a fair comparison with the state-of-the-art, we compute the AUROC jointly over all test frames (micro-averaging) without any video-level normalization. We follow the established evaluation protocol, applying temporal Gaussian 1D-smoothing on the aggregated anomaly scores \cite{delic2025seeker, Hirschorn_2023_ICCV, micorek24cvpr}. Additional results using Average Precision (AP) are provided in the supplementary material.

\textbf{Implementation Details.} To ensure a direct comparison to our closest competitors, we utilize the same extracted poses (via AlphaPose~\cite{alphapose}), raw confidences, and per-segment standardizations as the current state-of-the-art models, SeeKer~\cite{delic2025seeker} and STG-NF~\cite{Hirschorn_2023_ICCV}. Notably, this pose estimator introduces an 18-point skeleton. Each sequence is scored based on the current frame and the preceding $T-1$ frames. As illustrated in \Cref{fig:architecture}, our EBM is built with four hidden blocks, utilizing a hidden dimension of 1024.
To construct the multiscale DSM objective, we sample uniformly from a discrete geometric sequence of $L=10$ noise scales bounded between $\sigma_{low}=0.1$ and $\sigma_{high}=1.0$, unless stated otherwise.

The network is trained exclusively on normal data for up to 400 epochs with a batch size of 1024 using the AdamW optimizer (weight decay $10^{-2}$, $\beta_1=0.5$, $\beta_2=0.9$).
The learning rate follows a Cosine Annealing schedule with a linear warmup during the first epoch. The initial learning rate is set to $5 \times 10^{-4}$ for UBnormal and $2 \times 10^{-4}$ for ShanghaiTech, smoothly decaying to a minimum of half the initial rate. Because UBnormal provides a dedicated validation split, we employ early stopping based on validation AUROC; for ShanghaiTech, models are evaluated exactly at the end of the 400-epoch training schedule. Unless stated otherwise, we report AUROC for $T=12, K=48$ for both ShanghaiTech and UBnormal, selected based on the best performance on the UBnormal validation set. Finally, we maintain a secondary Exponential Moving Average (EMA) model, initialized as an exact copy of the actively trained model, whose weights are updated after each optimization step using a decay factor of 0.999. By averaging the weight updates over time, the EMA model smooths out the slight performance fluctuations observed in the active model during training, promoting a stable and highly reproducible model state. Consequently, this EMA model is strictly used for all our evaluations. To ensure rigorous evaluation against the state-of-the-art, we evaluate our main STEP framework across 20 independent training runs to report mean and standard deviation metrics.

\subsection{State-of-the-Art Comparison}
\Cref{tab:sota_comparison} compares our proposed STEP framework against existing state-of-the-art skeleton-based methodologies. First, we highlight the performance of our direct methodological baseline, MULDE~\cite{micorek24cvpr}. Surprisingly, adapting their score-matching framework to raw joint coordinates using only isolated, single-frame poses ($T=1$, lacking any temporal modeling) yields an AUROC of 80.6\% on UBnormal. This already surpasses previous complex temporal models like SeeKer (77.9\%). However, as we will show in \cref{sec:ablation}, this raw coordinate approach suffers from structural collapse when temporal windows are expanded. By projecting the poses into our whitened PC-space, STEP drastically resolves these limitations and unlocks the full potential of temporal density estimation.

On the synthetic UBnormal dataset, STEP achieves an average AUROC of 90.1\% ($\pm$0.4) across 20 independent training runs on the Full test set, and 90.9\% on the Human-Related split. This represents a substantial absolute improvement of over 12\% compared to the previous state-of-the-art, setting a new benchmark for the dataset. Furthermore, STEP demonstrates robust generalization on the real-world ShanghaiTech dataset. It achieves a leading average performance of 86.2\% ($\pm$0.1) on the Full set and 87.7\% on the HR split, on par with STG-NF (85.9\% and 87.4\%, respectively).

\begin{figure}[t]
  \centering
  \begin{tabular}{@{}c@{\hspace{2pt}}c@{}}
    \begin{tikzpicture}
      \node[anchor=south west, inner sep=0] (leftimg) at (0,0) {
        \includegraphics[width=0.5\textwidth]{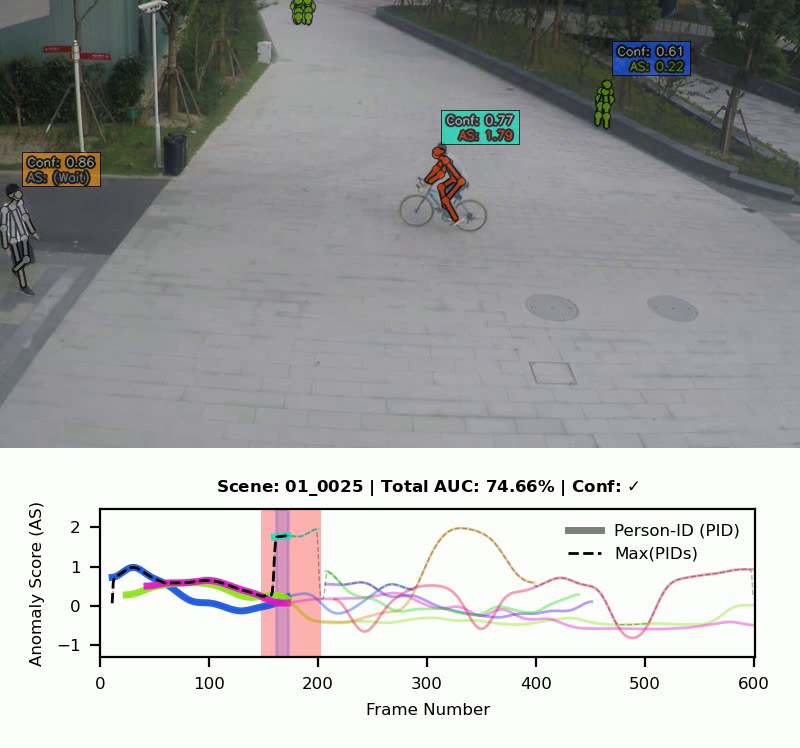}
      };
      \begin{scope}[x={(leftimg.south east)}, y={(leftimg.north west)}]
        \tikzset{bigarrow/.style={-{Latex[length=2mm, width=1mm]}, black, line width=0.25pt, opacity=0.3}}
        \draw[bigarrow] (0.355, 0.3) -- (0.52, 0.68);
      \end{scope}
    \end{tikzpicture} &
    \begin{tikzpicture}
      \node[anchor=south west, inner sep=0] (rightimg) at (0,0) {
        \includegraphics[width=0.5\textwidth]{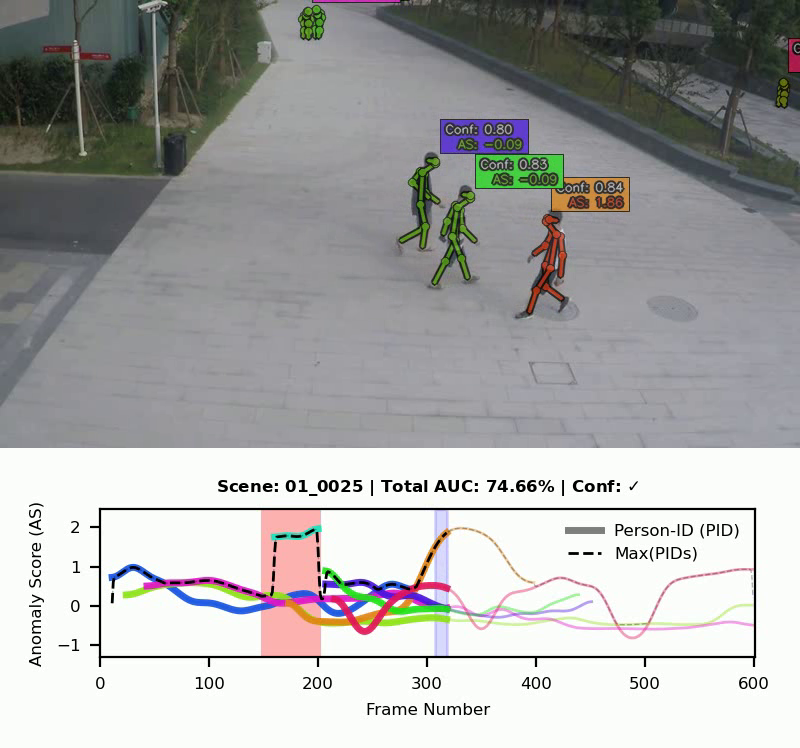}
      };
      \begin{scope}[x={(rightimg.south east)}, y={(rightimg.north west)}]
        \tikzset{bigarrow/.style={-{Latex[length=2mm, width=1mm]}, black, line width=0.25pt, opacity=0.3}}
        \draw[bigarrow] (0.55, 0.3) -- (0.65, 0.56);
      \end{scope}
    \end{tikzpicture}
  \end{tabular}
  \caption{\textbf{Qualitative results on ShanghaiTech (clip \texttt{01\_0025}).} Each person's track is assigned a unique color (its text box matches the time-series plot below each frame); skeleton joints are colored from green (low anomaly score) to red (high anomaly score). The number shown above each person is the mean sequence confidence; below it is the confidence-weighted anomaly score. The temporal plot shows all active tracks' scores over time; the red-shaded region marks the ground-truth anomaly interval.
  \textbf{Left,} frames 150--200: STEP correctly detects a passing cyclist (cyan track) as anomalous, producing a sharp anomaly score spike.
  \textbf{Right,} frames 300--370: An insightful false positive: a pedestrian (orange track) turns and walks backwards, receiving an elevated score despite no anomaly label, revealing the model has learned forward-facing walking as normal.}
  \label{fig:qualitative}
\end{figure}

Notably, our pose-only method outperforms several established methods that rely on heavier, multi-modal inputs (such as raw pixels and optical flow). A comprehensive comparison against these multi-modal approaches is provided in the supplementary material.
A qualitative illustration of STEP's detection behavior on ShanghaiTech is provided in \cref{fig:qualitative}. We further evaluate on the MSAD dataset~\cite{msad2024} and its Human-Related split (MSAD-HR), first explored for skeleton-based VAD by SeeKer~\cite{delic2025seeker}; STEP achieves 74.1\% AUROC on MSAD-HR, outperforming SeeKer (61.1\%) by 13\% and STG-NF (55.7\%) by 18\%. A detailed study on the MSAD dataset is provided in the supplementary material.

\begin{table}[t]
\setlength{\tabcolsep}{8pt}
\scriptsize
\centering
\caption{State-of-the-art comparison on ShanghaiTech and UBnormal. STEP sets a new state-of-the-art on UBnormal and matches performance on ShanghaiTech. Results are reported for both the Full test set and the Human-Related (HR) splits. To ensure rigorous evaluation, the STEP results are reported as the mean and standard deviation across 20 independent training runs. Best results per column are in \textbf{bold}, second best are \underline{underlined}. Results show AUROC (\%). $^\dagger$reproduced by us.}
\label{tab:sota_comparison}
\begin{tabular}{l ll ll}
\toprule
\multirow{2}{*}{Method} & \multicolumn{2}{l}{ShanghaiTech} & \multicolumn{2}{l}{UBnormal} \\
 & Full & HR & Full & HR \\
\midrule
BiPOCO \cite{miracle2022arxiv} & - & 74.9 & 50.7 & 52.3 \\
MPED-RNN \cite{morais2019cvpr} & 73.4 & 75.4 & 60.6 & 61.2 \\
MTP \cite{rodrigues2020wacv} & 76.0 & 77.0 & - & - \\
GEPC \cite{markovitz2020cvpr} & 76.1 & 74.8 & 53.4 & 55.2 \\
PoseCVAE \cite{jain2020icpr} & - & 75.5 & - & - \\
Normal Graph \cite{luo2020neurcomp} & - & 76.5 & - & - \\
COSKAD \cite{flaborea2024pr} & - & 77.1 & 65.0 & 65.5 \\
GCAE-LSTM \cite{li2021neurocmp} & - & 77.2 & - & - \\
MoCoDAD \cite{Flaborea_2023_ICCV} & - & 77.6 & 68.3 & 68.4 \\
MULDE (Pose, $T=1$) \cite{micorek24cvpr} & 78.5 & - & \underline{80.6}$^\dagger$ & - \\
STG-NF \cite{Hirschorn_2023_ICCV} & \underline{85.9} & \underline{87.4} & 71.8 & 71.5 \\
SeeKer \cite{delic2025seeker} & 85.5 & 86.9 & 77.9 & 78.9 \\
\midrule
\textbf{STEP (Ours)} & \textbf{86.2} $\pm$ 0.1 & \textbf{87.7} $\pm$ 0.1 & \textbf{90.1} $\pm$ 0.4 & \textbf{90.9} $\pm$ 0.4\\
\bottomrule
\end{tabular}
\end{table}

\subsection{Ablation Studies}
\label{sec:ablation}

\textbf{Resolving Structural Temporal Collapse.} Our core hypothesis is two-fold: (1) Denoising Score Matching fails on temporal pose sequences because isotropic coordinate noise generates physically impossible poses in high dimensions, and (2) off-the-shelf pose estimators introduce severe tracking artifacts that corrupt the learned energy. To cleanly isolate the relative impact of individual architectural choices, all ablation studies in this section are conducted using a fixed random seed. We note that our selected baseline configuration ($T=12, K=48$) achieves 90.4\% AUROC on UBnormal under this fixed random seed, tightly aligning with the highly stable multi-run mean (90.1\% $\pm$ 0.4) reported in \cref{tab:sota_comparison}.

\Cref{tab:dimensionality_ablation} systematically validates our core claims by isolating both mechanisms. When applying the EBM directly to raw coordinate vectors (MULDE baseline), performance peaks at an extremely short temporal window ($T=2$ at 80.7\%) but degrades rapidly, dropping by 10\% as the window expands to $T=32$. Integrating our sequence-level confidence weighting on raw coordinates provides a massive baseline improvement (boosting $T=2$ to 87.2\%), demonstrating that the EBM is highly sensitive to noisy pose estimates. However, confidence weighting alone cannot prevent the fundamental high-dimensional collapse, as performance still sharply declines to 73.1\% at $T=32$.

Projecting the sequences onto a compact PC-space without confidence weighting ($K=48, \times$) acts as a structural anchor. While absolute performance is bottlenecked by unmitigated tracking noise ($\sim$84\%), the model becomes more robust to temporal expansion, successfully halting the severe degradation seen in the raw coordinate baselines at high $T$.

By uniting the PCA approach with confidence weighting (Full STEP), we resolve both deficiencies. The table reveals a clear diagonal shift in performance based on the chosen capacity: overly tight bottlenecks ($K=16, 32$) perform exceptionally well for short windows ($T=4, T=2$) but lose critical information as sequence length increases. Conversely, under-compressing the space ($K=128$) leaves the EBM susceptible to the dimensionality curse at short windows. While multiple balanced configurations ($K \in \{48, 64\}$) achieve $>90\%$ AUROC at longer temporal windows, we anchor our final evaluation to the $T=12, K=48$ configuration (yielding 90.4\% on the test set), as it achieved the highest performance on the UBnormal validation set.

\begin{table}[t]
\setlength{\tabcolsep}{6pt}
\scriptsize
\centering
\caption{Ablation of temporal window ($T$), PCA dimension ($K$), and confidence weighting on UBnormal. Naively applying DSM to raw coordinates (MULDE) causes structural collapse as $T$ expands. Projecting sequences onto our PC-space stabilizes this degradation, while confidence weighting mitigates tracking noise. The combination reveals a clear capacity trade-off, with the highest-performing configurations shifting diagonally as $T$ increases. Results show AUROC (\%). Best results per row are in \textbf{bold}, second best are \underline{underlined}.
Missing values (-) denote invalid configurations where the target PCA dimension $K$ exceeds the input dimension. All results are reported for a fixed random seed. $^\dagger$reproduced by us.}
\label{tab:dimensionality_ablation}
\begin{tabular}{@{}l c c | cccccccc@{}}
\toprule
\multirow{2}{*}{Method} & \multirow{2}{*}{PCA ($K$)} & \multirow{2}{*}{Conf.} & \multicolumn{8}{c}{Temporal Window ($T$)} \\
 & & & 1 & 2 & 4 & 8 & 12 & 16 & 24 & 32 \\ \midrule
MULDE$^\dagger$ & $\times$ & $\times$ & \underline{80.6} & \textbf{80.7} & 78.7 & 75.7 & 72.2 & 72.6 & 72.6 & 70.6 \\
STEP & $\times$ & \checkmark & 86.0 & \textbf{87.2} & \underline{86.3} & 82.4 & 81.3 & 80.5 & 78.3 & 73.1 \\
STEP & 48 & $\times$ & - & 82.1 & 83.2 & \textbf{84.1} & \underline{84.0} & 83.6 & 82.3 & 81.3 \\
\midrule
STEP & 16 & \checkmark & 86.7 & \underline{87.9} & \textbf{88.0} & 87.3 & 86.4 & 85.0 & 82.9 & 81.6 \\
STEP & 32 & \checkmark & 86.7 & \textbf{89.5} & 88.4 & 88.6 & 88.9 & \underline{89.0} & 87.5 & 85.7 \\
STEP& 48 & \checkmark & - & 88.8 & 89.6 & 89.2 & \textbf{90.4} & \underline{90.2} & 89.5 & 88.3 \\
STEP & 64 & \checkmark & - & 88.2 & 88.8 & 90.0 & 90.1 & \textbf{90.5} & \underline{90.3} & 88.8 \\
STEP & 128 & \checkmark & - & - & 87.5 & 87.9 & 88.8 & \textbf{89.2} & \textbf{89.2} & 88.4 \\
\bottomrule
\end{tabular}
\end{table}

\textbf{Component Breakdown and Noise Scale Sensitivity.} To isolate the impact of our specific design choices, we ablate the core components of the STEP framework in \cref{tab:unified_ablation}. By reporting the degradation across our unified configuration ($T=12, K=48$) on both datasets, we demonstrate that these trends are fundamental to the framework. A complete ablation matrix detailing every $T$ and $K$ combination is provided in the supplementary material.

First, we observe that our $\sigma$-modulated Residual MLP consistently outperforms the standard unmodulated MLP used in previous EBM frameworks by roughly 1\%, suggesting the network requires stronger architectural conditioning to utilize the multi-scale noise. Furthermore, standardizing (whitening) the PC-space is critical; without it, the isotropic noise $\mathcal{N}(0, \sigma^2 I)$ disproportionately corrupts low-variance principal components while under-perturbing the dominant ones, causing performance to drop by 2.2\% on UBnormal.

Next, we evaluate the role of pose estimator confidence. A common heuristic, \eg used by SeeKer~\cite{delic2025seeker}, is to drop poses below a certain confidence threshold (\eg, $<0.4$) during training. While applying a highly conservative filter ($<0.1$) yields a marginal 0.1\% improvement for our model, relying on such thresholds is inherently brittle and introduces another hyperparameter. Aggressive filtering ($<0.4$) discards too much valid, albeit noisy, training data, causing a massive 4.5\% performance collapse on UBnormal. By dropping these low-confidence poses, the model fails to learn the natural variance of fast or occluded normal motions, creating blind spots in the learned density. Instead, our soft, sequence-level confidence weighting dynamically scales the loss. This allows the model to leverage the full training set while naturally down-weighting tracking failures.

Finally, we evaluate the bounds of the noise scales ($\sigma_{low}=0.1, \sigma_{high}=1.0$) employed in the DSM objective. As shown in the table, the model is robust within a reasonable range (\eg, lowering $\sigma_{high}$ to 0.5 yields comparable results), but performance degrades at the extremes. Setting the minimum noise scale too low ($\sigma_{low}=10^{-4}$) causes the network to become overly sensitive to the microscopic measurement noise of the pose estimator rather than evaluating the macroscopic validity of the pose itself. By maintaining a lower bound ($\sigma_{low}=0.1$), the noise acts as an implicit kernel smoother in the PC space, absorbing minor coordinate fluctuations and forcing the network to model true kinematic structure rather than memorizing noisy training samples. Conversely, if $\sigma_{low}$ is too large (0.5), the density estimate becomes over-smoothed, blurring the fine-grained, high-density basins of normal motion and dropping performance by 3.5\% on UBnormal and 3.6\% on ShanghaiTech. At the other end of the spectrum, an insufficient maximum noise scale ($\sigma_{high}=0.2$) limits the spatial reach of the score estimator. While this narrower coverage is adequate for the subtle anomalies in ShanghaiTech, it fails to provide gradient signals in the sparse regions distant from the primary PC-space cluster. Consequently, the network remains untrained for the extreme anomalous poses found in UBnormal, resulting in less reliable energy scores and a slight performance drop (-0.5\%). Our selected bounds provide adequate density coverage for the unit-variance PCA space across both benchmarks.

\begin{table}[tb]
\setlength{\tabcolsep}{10pt}
\scriptsize
\centering
\caption{Component Ablation of STEP across both datasets at $T=12, K=48$. Removing our core architectural choices or aggressively dropping uncertain training poses results in consistent performance drops across both datasets. All AUROC (\%) results are reported for a fixed seed.}
\label{tab:unified_ablation}
\begin{tabular}{@{}l r r r r@{}}
\toprule
Configuration & \multicolumn{2}{c}{ShanghaiTech} & \multicolumn{2}{c}{UBnormal}  \\
\midrule
\textbf{Full STEP (Ours)} & \textbf{86.3} & $\Delta$ & \textbf{90.4} & $\Delta$ \\ \midrule
\textit{Architecture \& PC-space} & & & & \\
\quad w/o $\sigma-$modulation Residual MLP (Vanilla MLP) & 85.6 & \textcolor{red}{-0.7} & 89.3 & \textcolor{red}{-1.1} \\
\quad w/o Whitening & 85.5 & \textcolor{red}{-0.8} & 88.2 & \textcolor{red}{-2.2} \\ \midrule
\textit{Training Confidence Handling} & & & & \\
\quad Drop Pose (Confidence $< 0.1$) & 86.4 & +0.1 & 90.5 & +0.1 \\
\quad Drop Pose (Confidence $< 0.2$) & 86.0 & \textcolor{red}{-0.3} & 89.6 & \textcolor{red}{-0.8} \\
\quad Drop Pose (Confidence $< 0.4$) & 84.3 & \textcolor{red}{-2.0} & 85.9 & \textcolor{red}{-4.5} \\ \midrule
\textit{Noise Boundaries (vs. $\sigma_{low}=0.1, \sigma_{high}=1.0$)} & & & & \\
\quad Minimum Noise $\sigma_{low} = 10^{-4}$ & 86.1 & \textcolor{red}{-0.2} & 89.6 & \textcolor{red}{-0.8} \\
\quad Minimum Noise $\sigma_{low} = 0.5$ & 82.7 & \textcolor{red}{-3.6} & 86.9 & \textcolor{red}{-3.5} \\ \midrule
\quad Maximum Noise $\sigma_{high} = 0.2$ & 86.4 & +0.1 & 89.9 & \textcolor{red}{-0.5} \\
\quad Maximum Noise $\sigma_{high} = 0.5$ & 86.4 & +0.1 & 90.6 & +0.2 \\
\bottomrule
\end{tabular}
\end{table}

\subsection{Computational Efficiency}
\label{subsec:efficiency}
Our lightweight architecture is highly efficient. Timing the complete pipeline, from the already-tracked pose sequences through PCA projection, whitening, and final anomaly scoring, reveals that latency is practically invariant to $T$. On a NVIDIA GTX 1080, this entire scoring process for 50 persons per frame takes $<1$\,ms ($\sim$1026\,FPS) with a 32.6\,MB memory footprint. Scaling to 100 persons takes just 1.62\,ms ($\sim$618\,FPS, 44.9\,MB). Thus, STEP provides real-time, state-of-the-art anomaly detection at a fractional cost to the underlying pose extraction.

%% file: 05_discussion.tex
\section{Limitations}
\label{sec:discussion}
While our proposed STEP framework sets a new benchmark for skeleton-based VAD, we acknowledge several important limitations regarding our design choices.

\textbf{Pose Estimator:} STEP relies on the upstream pose estimators. While total occlusion inevitably interrupts data extraction, the temporal nature of our sequence-level scoring ensures that the irregular kinematics leading up to an anomaly (\eg, falling) are typically flagged before the subject is fully obscured.

\textbf{Contextual and Multi-Person Anomalies:} By design, our framework evaluates subjects independently and strips away pixel-level appearance to ensure robustness against background clutter and to preserve privacy. However, a fundamental trade-off of this pure pose formulation is that it naturally struggles with anomalies defined entirely by multi-person interactions (\eg, illegal exchange of goods) or object manipulation (\eg, abandoning a bag) if the underlying posture remains typical. We argue that while reintroducing visual appearance features could easily capture this missing context, doing so fundamentally compromises the privacy-preserving properties that motivate skeleton-based anomaly detection in the first place. Instead, explicitly modeling multi-person relational dynamics using purely pose features remains a key direction for future work.

%% file: 06_conclusion.tex
\section{Conclusion}
\label{sec:conclusion}

We presented STEP, a robust and highly efficient score-based framework for skeleton-based Video Anomaly Detection. To prevent the structural collapse that occurs when applying Denoising Score Matching directly to raw temporal coordinates, we project the pose sequences onto a compact, whitened PC-space. This low-dimensional representation, coupled with a soft sequence-level confidence weighting, allows our lightweight model to robustly score anomalies while naturally mitigating upstream tracking artifacts. Consequently, our architecture is efficient, capable of processing dense crowds with minimal computational overhead. Extensive experiments confirm the effectiveness of our approach, with STEP setting a new state-of-the-art on UBnormal and matching the performance on ShanghaiTech.

%% file: 07_acknowledgements.tex
\section*{Acknowledgements}
This work received funding from the BRIDGE programme of the Republic of Austria, Federal Ministry of Innovation, Mobility and Infrastructure (BMIMI) under the project ReCoDi (914991). The Austrian Research Promotion Agency (FFG) has been authorised for the programme management.

%% file: supplementary.tex
\section{Qualitative Examples}
\input{supp_01_qualitative}

\clearpage
\section{In-Depth Analysis of STEP Mechanics}
\input{supp_02_evaluation_TK_architecture}

\clearpage
\section{Extended Experimental Results}
\input{supp_03_evaluation_comprehensive}

\clearpage
\input{supp_04_pca}

%% file: supp_01_qualitative.tex
\label{supp:qualitative}

In both the provided videos and the static examples below, we utilize a consistent visualization scheme to overlay our model's predictions:
\begin{itemize}
    \item \textbf{Skeleton Color:} The skeletal joints and bones are color-coded based on the predicted anomaly score. Green indicates normal kinematic behavior, smoothly transitioning to red as the sequence is getting a higher anomaly score.
    \item \textbf{On-Screen Metrics:} The text box above each person displays two key metrics. The top value represents the mean pose confidence score ($c$) over the current temporal window $T$. The bottom value displays the final, confidence-weighted, scaled anomaly score. The color of the box uniquely identifies the person's track ID throughout the sequence.
    \item \textbf{Temporal Plot:} The graph situated below the video frame visualizes the anomaly scores of all active tracks over time. The line colors correspond directly to the text box color track IDs, allowing for easy tracking of an individual's score trajectory. The red background indicates the ground truth anomaly.
\end{itemize}

The following static figures highlight specific scenarios, demonstrating successful anomaly detections, the critical role of our confidence weighting mechanism in mitigating false positives during low-confident detections (\eg occlusions), and typical failure cases caused by extreme tracking errors.

\begin{figure}[thbp]
  \centering
  \begin{tikzpicture}
    \node[anchor=south west, inner sep=0] (myimage) at (0,0) {
      \begin{tabular}{@{}c@{}c@{}}
        \includegraphics[width=0.5\textwidth]{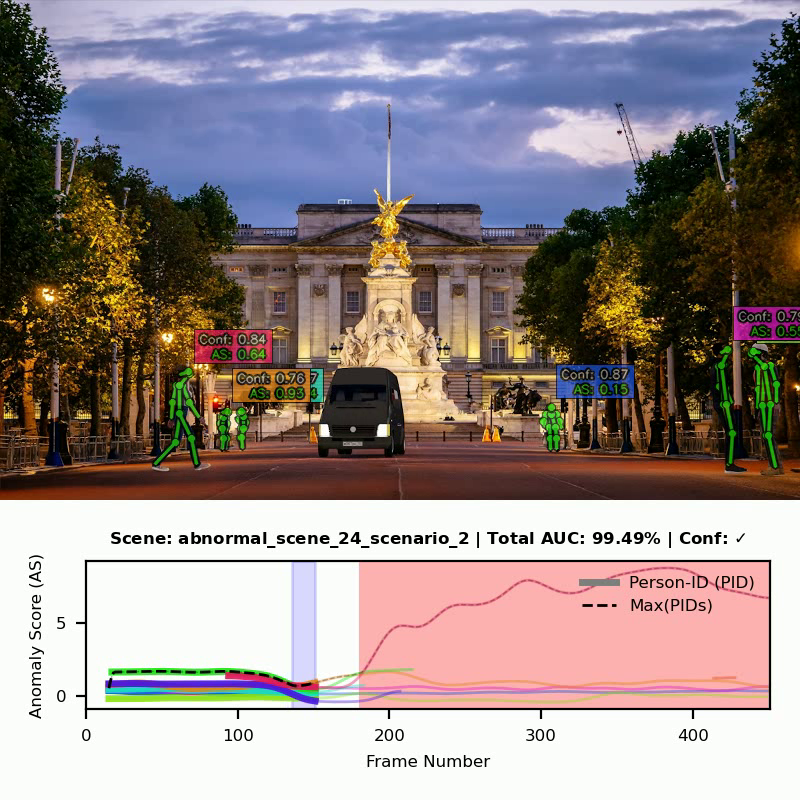} & 
        \includegraphics[width=0.5\textwidth]{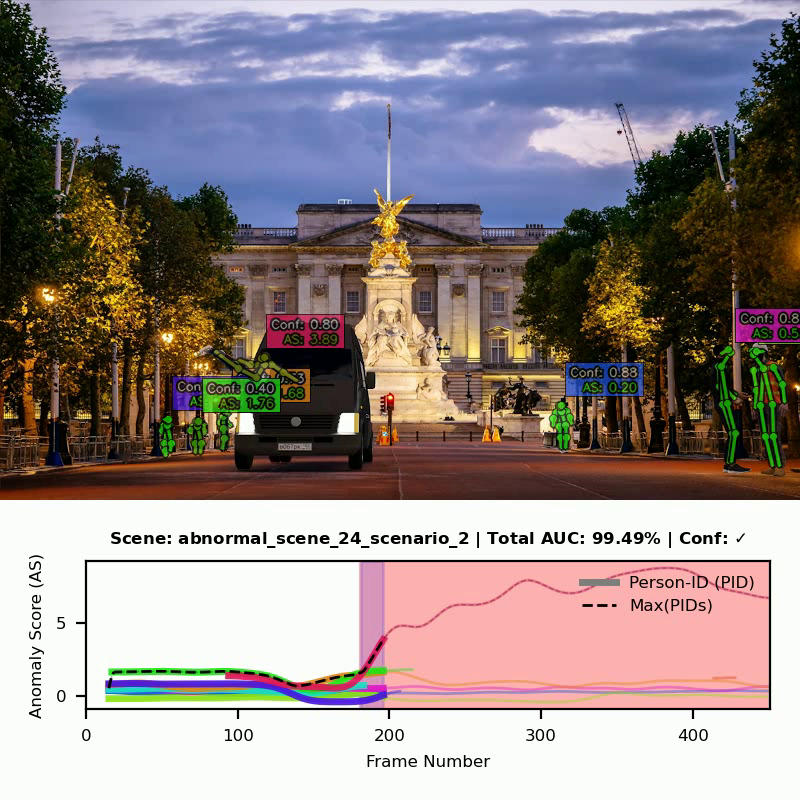} \\
        \includegraphics[width=0.5\textwidth]{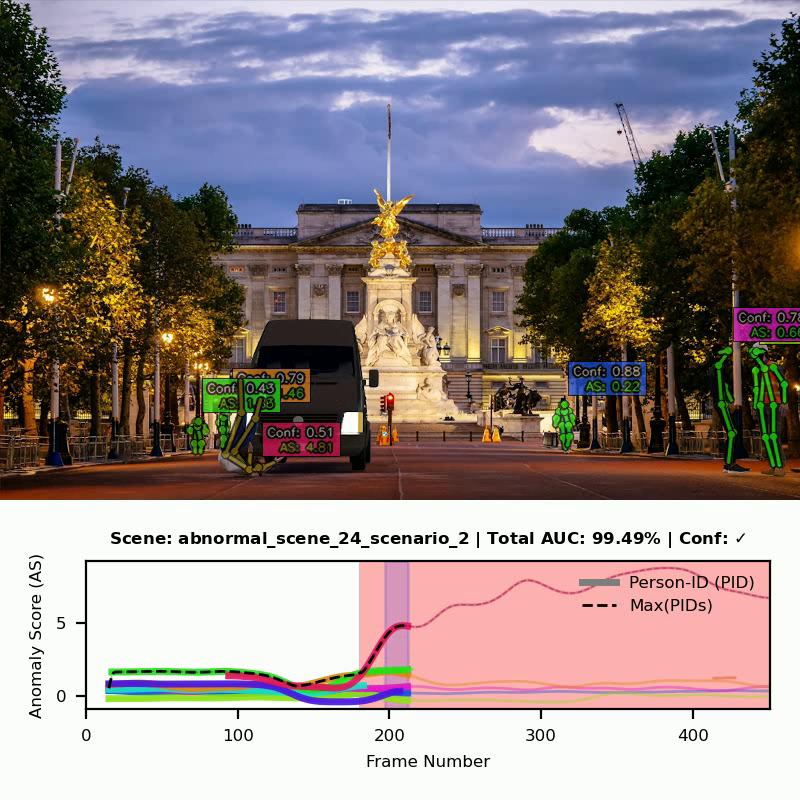} & 
        \includegraphics[width=0.5\textwidth]{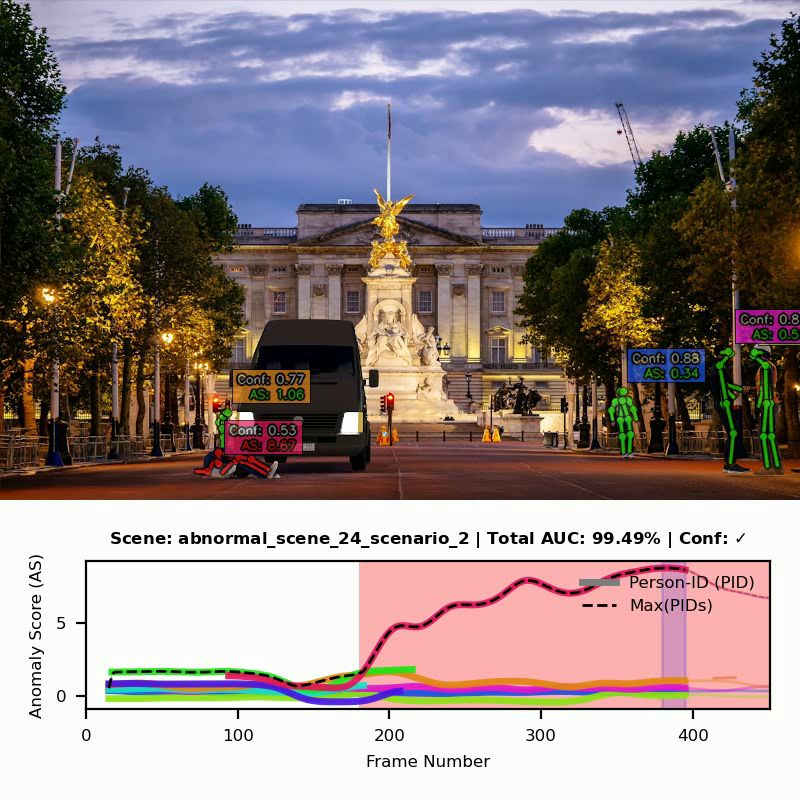}
      \end{tabular}
    };
  \end{tikzpicture}
  \caption{\textbf{Successful anomaly detection on UBnormal.} This four-frame sequence captures an extreme anomaly where a van strikes a pedestrian (identified by the red box). Our STEP framework accurately detects the severe kinematic violation: the anomaly score sharply spikes upon the initial impact and remains heavily elevated while the person is lying unnaturally on the ground. Both the dynamic fall and the static abnormal pose are correctly flagged as highly anomalous (indicated by the red skeletal overlay). Video clip: \texttt{abnormal\_scene\_24\_scenario\_2}.}
  \label{fig:qualitative_success}
\end{figure}

\begin{figure}[thbp]
  \centering
  \begin{tikzpicture}
    \node[anchor=south west, inner sep=0] (myimage) at (0,0) {
      \begin{tabular}{@{}c@{}c@{}}
        \includegraphics[width=0.5\textwidth]{01_0025_1.png} & 
        \includegraphics[width=0.5\textwidth]{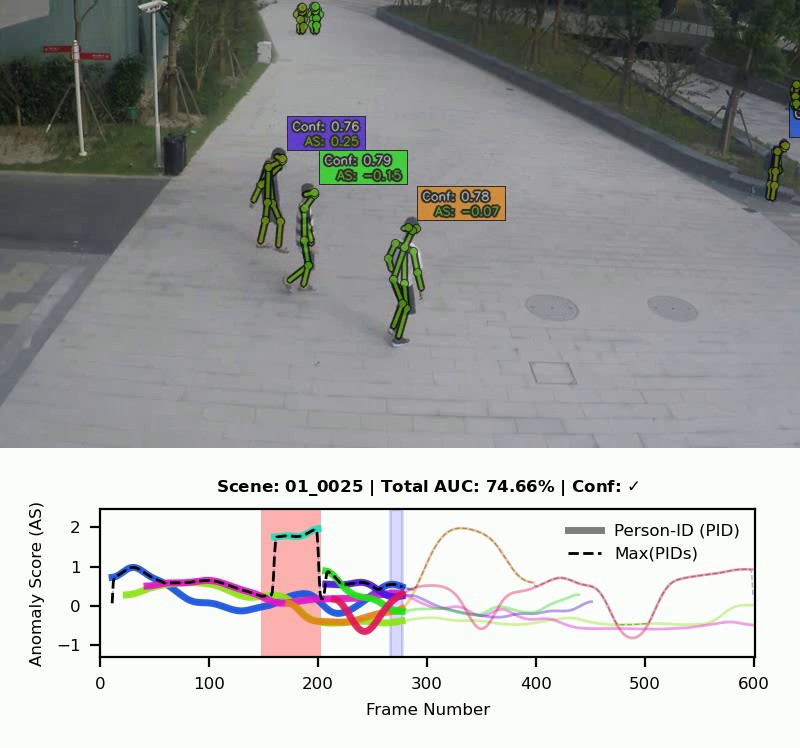} \\
        \includegraphics[width=0.5\textwidth]{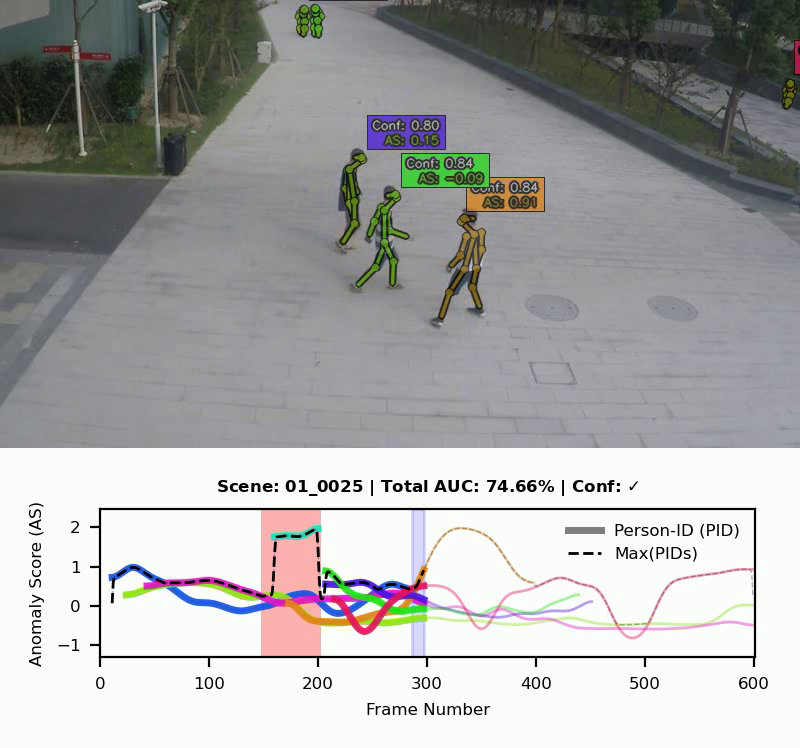} & 
        \includegraphics[width=0.5\textwidth]{01_0025_4.png}
      \end{tabular}
    };
  \end{tikzpicture}
  \caption{\textbf{Qualitative example on ShanghaiTech.} This four-frame sequence captures both a successful detection and an insightful false positive. First, our model successfully detects a passing cyclist (\ie cyan ID) as a clear anomaly. In the subsequent three frames, three pedestrians walk from left to right. One pedestrian (\ie orange ID) turns and proceeds to walk backwards. While backward walking is not labeled as an anomaly in the dataset ground truth, our system flags it with an elevated anomaly score. This highlights that our model effectively learned the standard, forward-facing pedestrian kinematics that define normality in the training set, causing it to intuitively recognize this rare backward motion as an anomalous event. Video clip: \texttt{01\_0025}.}
  \label{fig:qualitative_success_backward_walking}
\end{figure}

\begin{figure}[thbp]
  \centering
  \begin{tikzpicture}
    \node[anchor=south west, inner sep=0] (myimage) at (0,0) {
      \begin{tabular}{@{}c@{}c@{}}
        \includegraphics[width=0.5\textwidth]{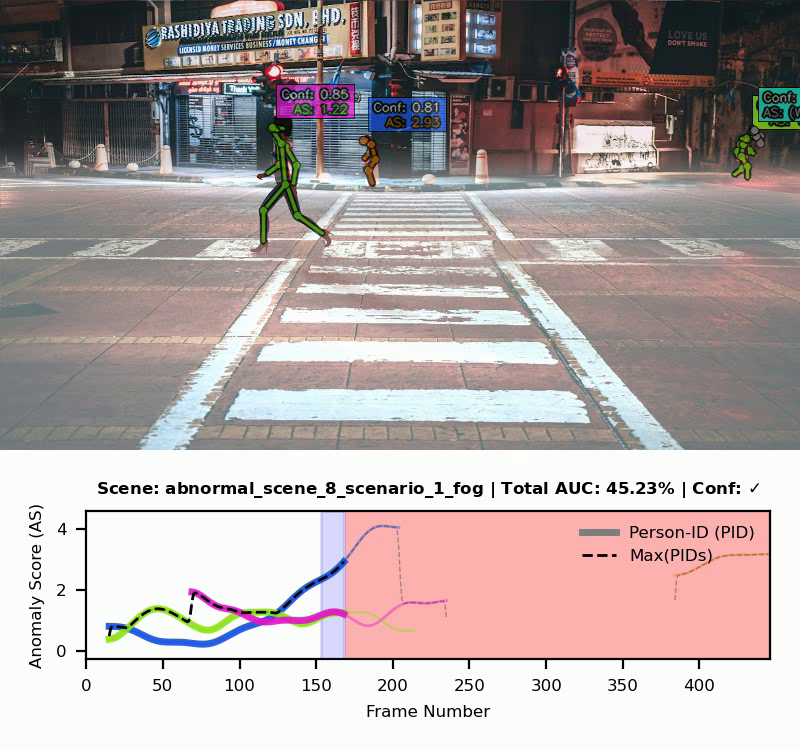} & 
        \includegraphics[width=0.5\textwidth]{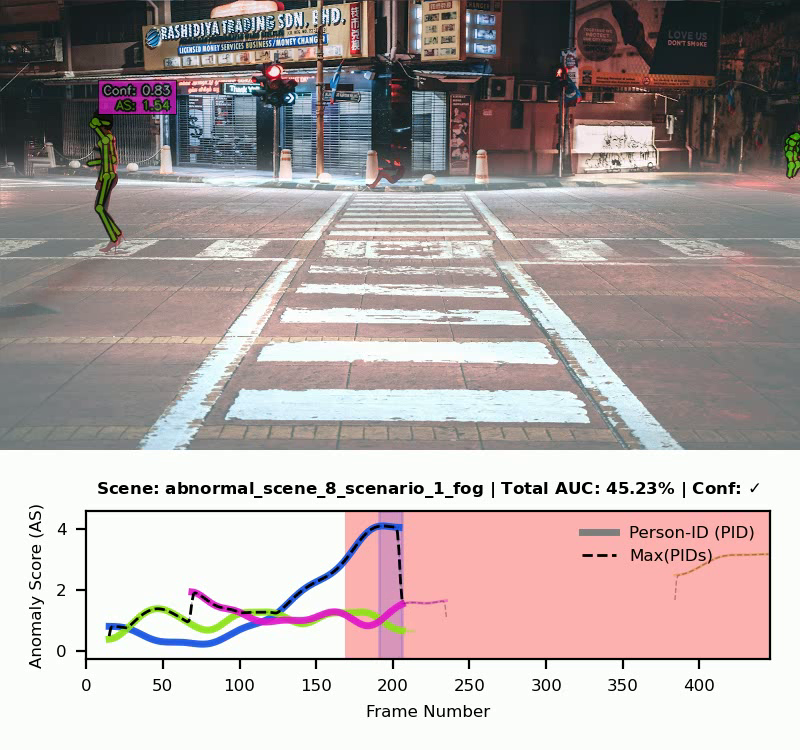} \\
        \includegraphics[width=0.5\textwidth]{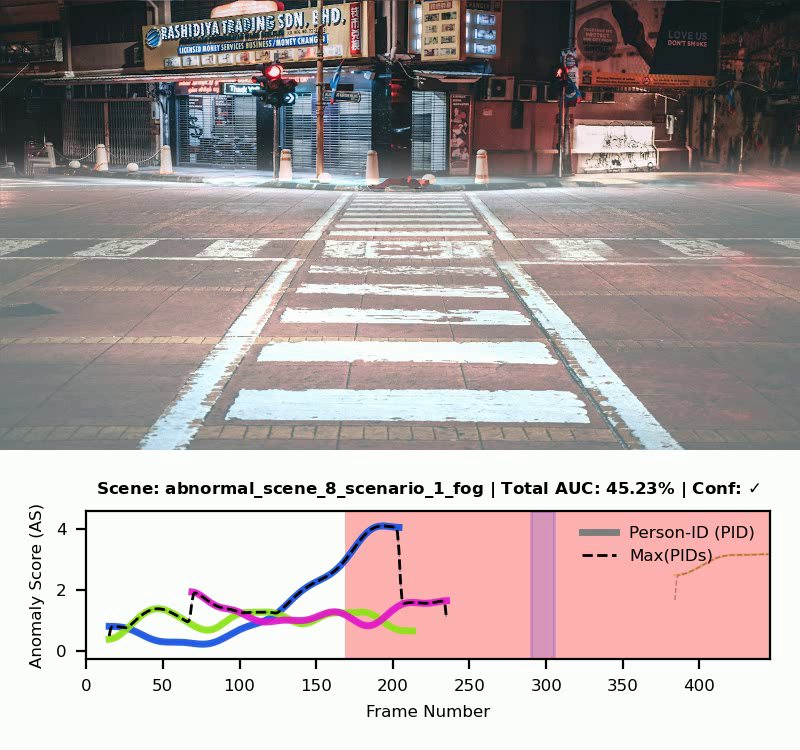} & 
        \includegraphics[width=0.5\textwidth]{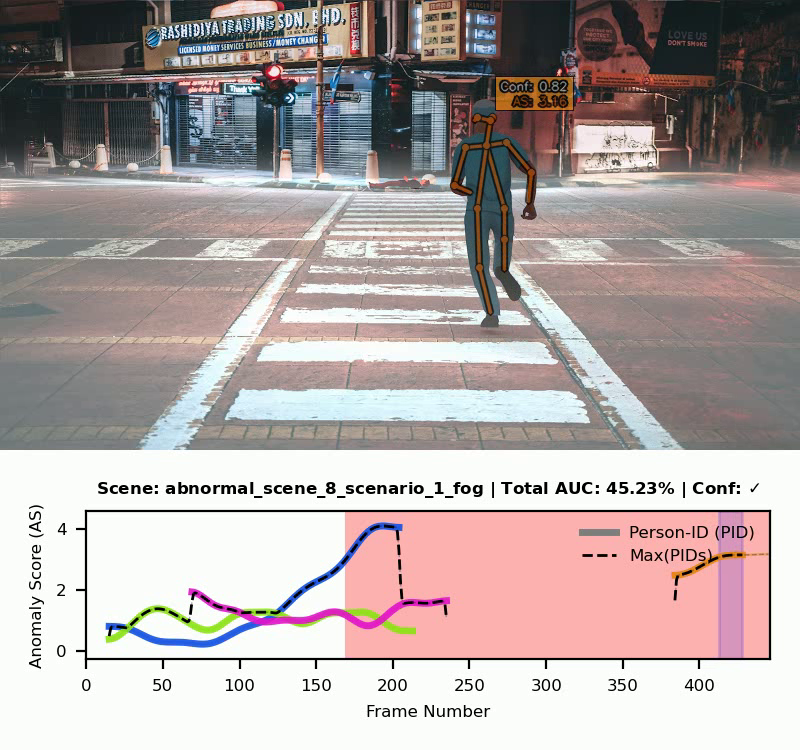}
      \end{tabular}
    };
  \end{tikzpicture}
  \caption{\textbf{Tracking failure and subsequent detection on UBnormal.} This sequence highlights a fundamental limitation of skeleton-based approaches: the reliance on upstream pose estimators. Initially, a pedestrian (\ie blue ID) experiences a seizure and collapses. However, due to the extreme pose and environmental factors (\eg, fog), the tracker completely loses the subject mid-fall. Because no skeleton is detected, our framework cannot evaluate the motion. Later in the sequence, a doctor (\ie orange ID) rushes to assist the fallen individual. The tracker successfully captures this new subject, and our system correctly flags the fast-paced running motion as anomalous. This demonstrates that while STEP is robust, its performance is ultimately bottlenecked by the reliability of the foundational pose extraction. Video clip: \texttt{abnormal\_scene\_8\_scenario\_1\_fog}.}
  \label{fig:qualitative_tracking_failure}
\end{figure}

\begin{figure}[thbp]
  \centering
  \begin{tikzpicture}
    \node[anchor=south west, inner sep=0] (myimage) at (0,0) {
      \begin{tabular}{c}
        \includegraphics[width=\textwidth]{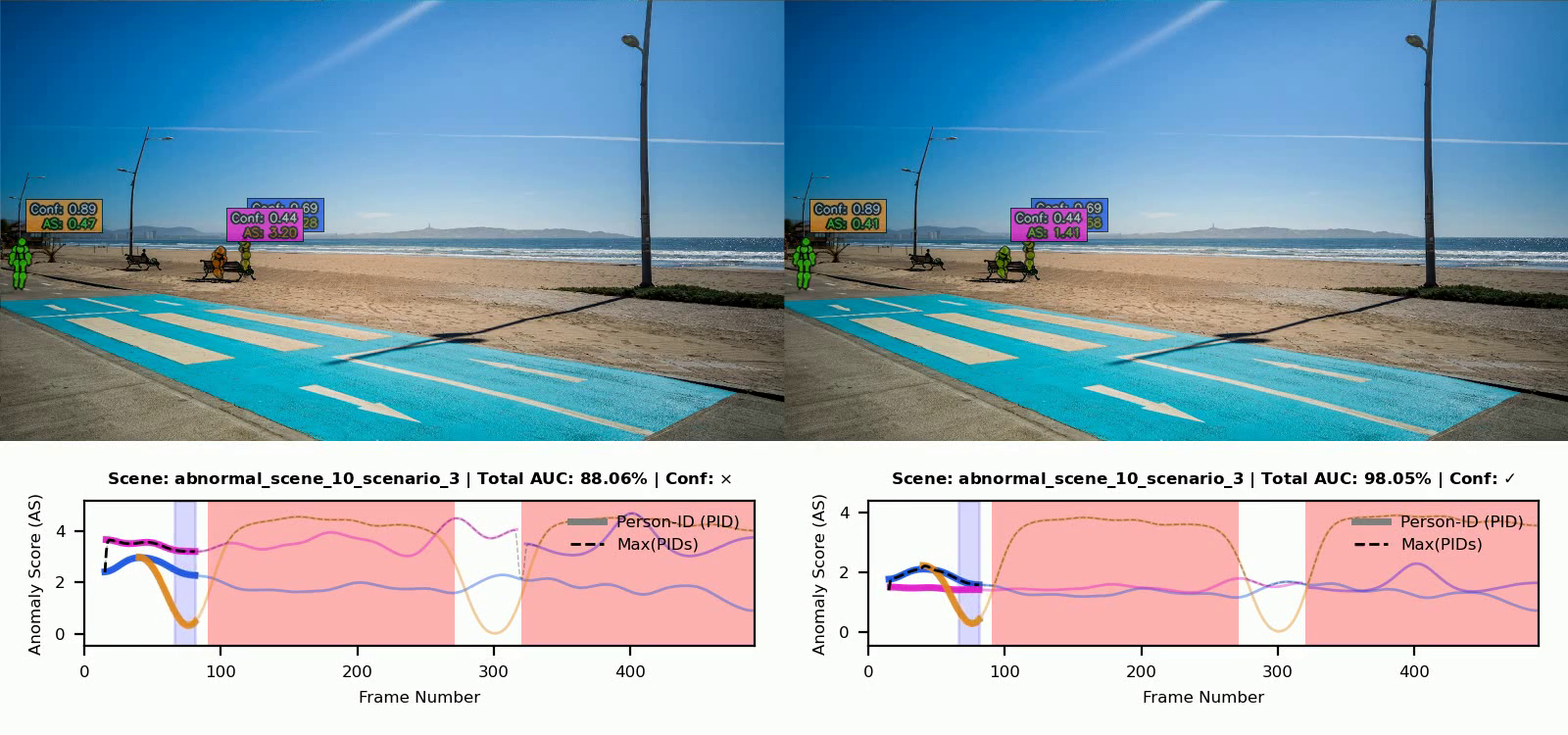} \\
        \includegraphics[width=\textwidth]{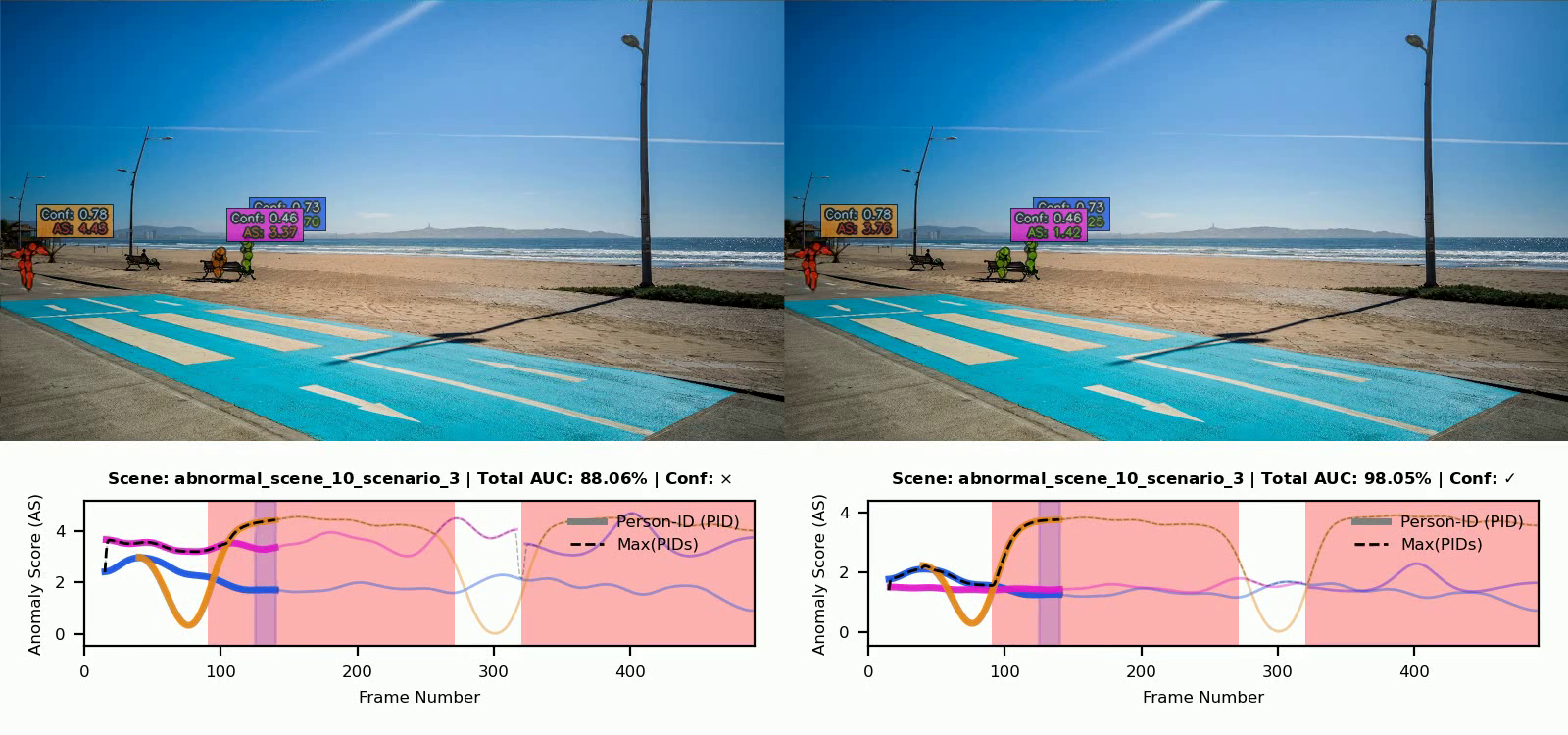} \\
        \includegraphics[width=\textwidth]{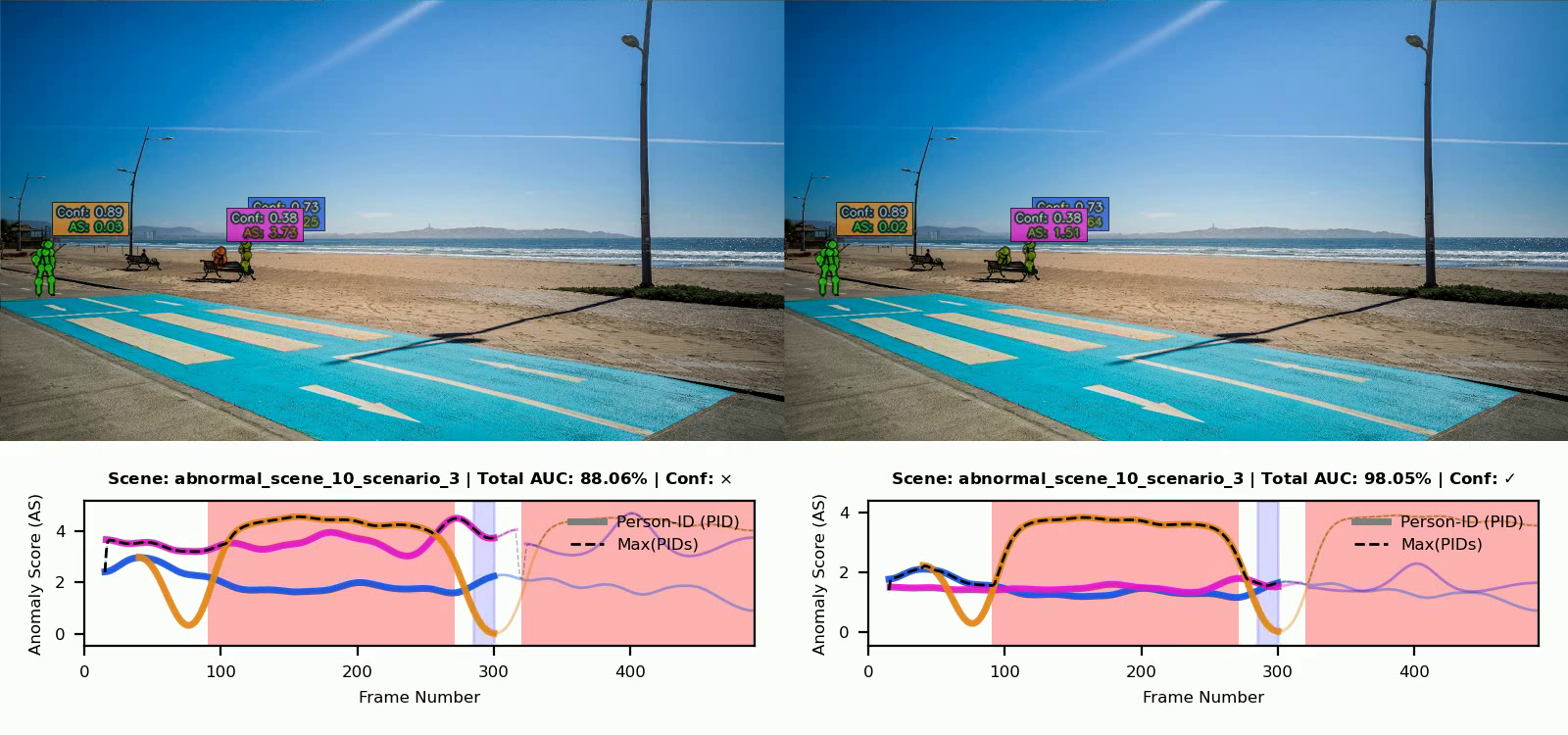}
      \end{tabular}
    };
    \begin{scope}[x={(myimage.south east)}, y={(myimage.north west)}]
      \tikzset{bigarrow/.style={-{Latex[length=2mm, width=1mm]}, black, line width=0.25pt}}
      \tikzset{highlight/.style={black, line width=0.3pt}}
      \draw[bigarrow] (0.125, 0.725) -- (0.05, 0.89);
      \draw[bigarrow] (0.125, 0.76) -- (0.17, 0.89);
      \draw[bigarrow] (0.625, 0.745) -- (0.675, 0.89);
      \draw[bigarrow] (0.17, 0.44) -- (0.05, 0.55);
      \draw[bigarrow] (0.315, 0.099) -- (0.18, 0.225);
      \draw[bigarrow] (0.31+0.5,0.075) -- (0.18+0.5, 0.225);
        \draw[highlight] (0.68, 0.90) ellipse (0.015 and 0.01);
    \end{scope}
  \end{tikzpicture}
  \caption{Confidence Weighting on UBnormal: We show the impact of using the confidence weighting. Left column without, right column with confidence weighting. The occluded person sitting on the bench has lower confidence. Without confidence reweighting, the normal sitting person would be wrongly detected as an anomaly. Video clip: \texttt{abnormal\_scene\_10\_scenario\_3}.}
  \label{fig:qualitative_confidence}
\end{figure}

Specifically, \cref{fig:qualitative_success} illustrates a highly successful detection on UBnormal, capturing both the dynamic motion of a pedestrian being struck and the subsequent static anomaly of the person lying on the ground. \cref{fig:qualitative_success_backward_walking} showcases our model's deep understanding of standard human kinematics on ShanghaiTech, successfully flagging a cyclist as well as an out-of-distribution backward-walking pedestrian. In \cref{fig:qualitative_tracking_failure}, we examine a fundamental limitation: a temporary false negative where environmental factors (\eg, fog) cause the upstream pose estimator to completely lose track of a collapsing pedestrian, though the system quickly recovers to detect a subsequent running anomaly. Finally, \cref{fig:qualitative_confidence} directly visualizes the necessity of our confidence weighting mechanism, proving how it actively suppresses false positives caused by noisy, low-confidence pose extractions of partially occluded individuals.

%% file: supp_02_evaluation_TK_architecture.tex
In this section, we provide a detailed visual analysis of the core mechanics driving the STEP framework. First, we explore the intricate relationship between temporal segment lengths ($T$) and PCA bottleneck capacities ($K$), demonstrating how our approach unlocks stable anomaly detection across extended temporal horizons where traditional raw-coordinate methods fail. Subsequently, we dissect the impact of our architectural design choices, offering an empirical sweep that proves our $\sigma$-modulated Residual MLP consistently yields better anomaly detection performance compared to standard network architectures.

\subsection{Temporal Window ($T$) and PCA Capacity ($K$)}
\begin{figure}[!thbp]
  \centering
  
  \begin{minipage}[c]{0.4\textwidth}
    \includegraphics[width=\textwidth]{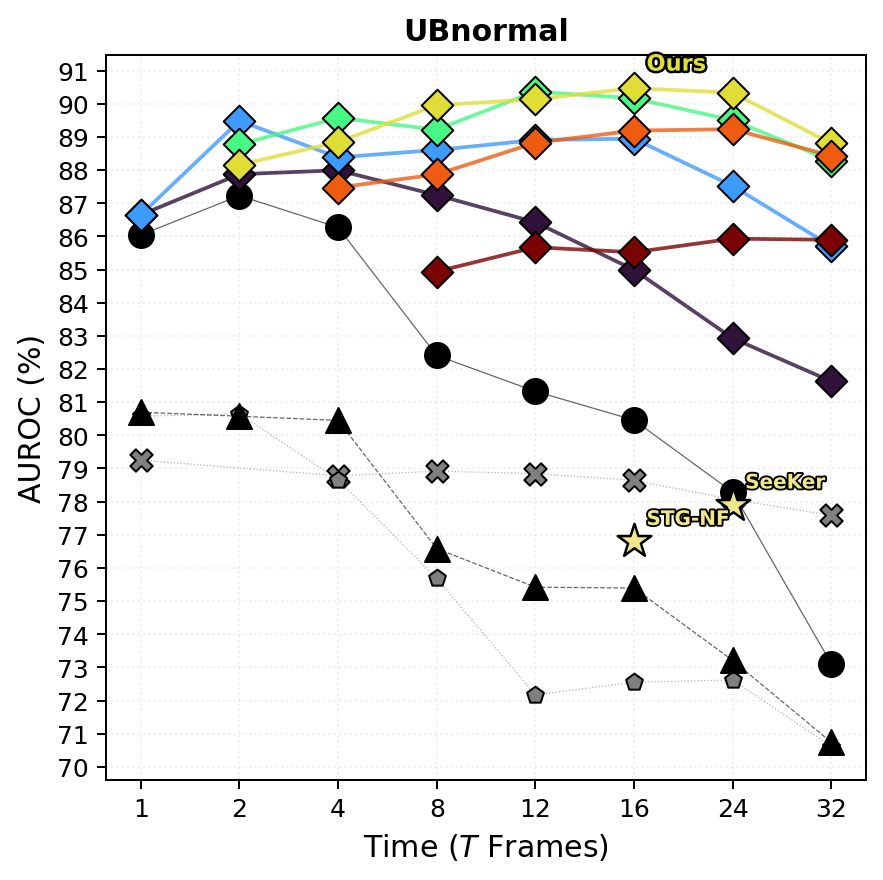}
  \end{minipage}\hfill
  \begin{minipage}[c]{0.4\textwidth}
    \includegraphics[width=\textwidth]{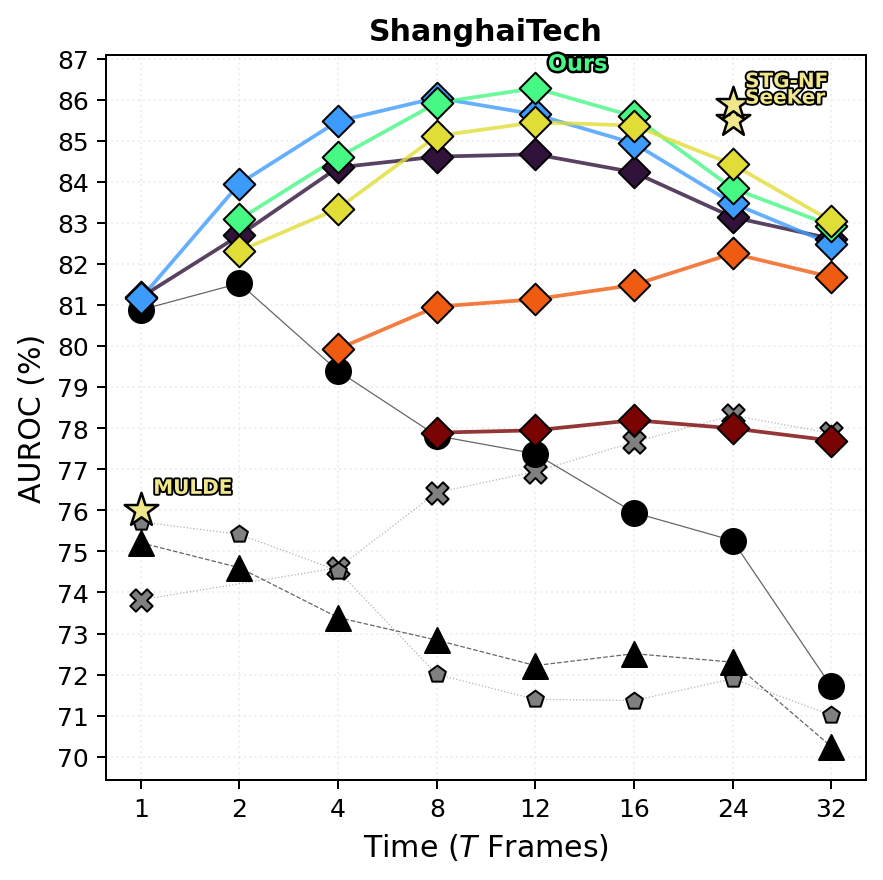}
  \end{minipage}\hfill
  \begin{minipage}[c]{0.2\textwidth}

    \includegraphics[scale=0.4]{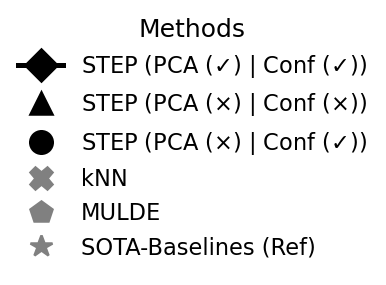}

    \vspace{5pt}

    \includegraphics[scale=0.4]{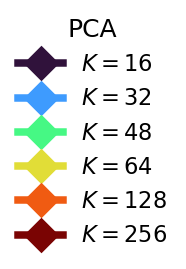}

  \end{minipage}

  \caption{\textbf{Exhaustive temporal and dimensional sweep.} We evaluate the impact of the temporal window length ($T$) and manifold dimensionality ($K$) on \textbf{UBnormal} (left) and  \textbf{ShanghaiTech} (right). Across both datasets, our PCA-projected STEP framework maintains stable, high performance at extended temporal window lengths, whereas the performance of baselines operating directly on raw coordinates degrades substantially as $T$ increases.}
  \label{fig:sweep_combined}
\end{figure}

In \cref{fig:sweep_combined}, we present a comprehensive performance sweep comparing the baseline MULDE (raw coordinates), a simple non-parametric kNN ($k=1$) baseline, and our STEP framework across varying temporal segment lengths ($T$) and PCA capacities ($K$). Expanding upon \cref{tab:dimensionality_ablation} of the main manuscript, these figures visually demonstrate the distinct behaviors of each method as the temporal horizon expands. Notably, while the parametric raw-coordinate baseline (MULDE) suffers from severe structural degradation as $T$ increases (evidenced by the sharp downward slope), the kNN baseline exhibits dataset-dependent behavior, actually improving with longer temporal windows on ShanghaiTech. Nevertheless, by projecting the motion into a compact PC-space, STEP effectively mitigates the degradation associated with high-dimensional inputs. It maintains highly stable performance across all evaluated temporal lengths, consistently outperforming both raw-coordinate approaches within a proper operating range of $K$.

\subsection{Architecture: $\sigma$-modulated Residual MLP vs. Vanilla MLP}
\begin{figure}[!thbp]
  \centering

  \begin{minipage}[c]{0.4\textwidth}
    \includegraphics[width=\textwidth]{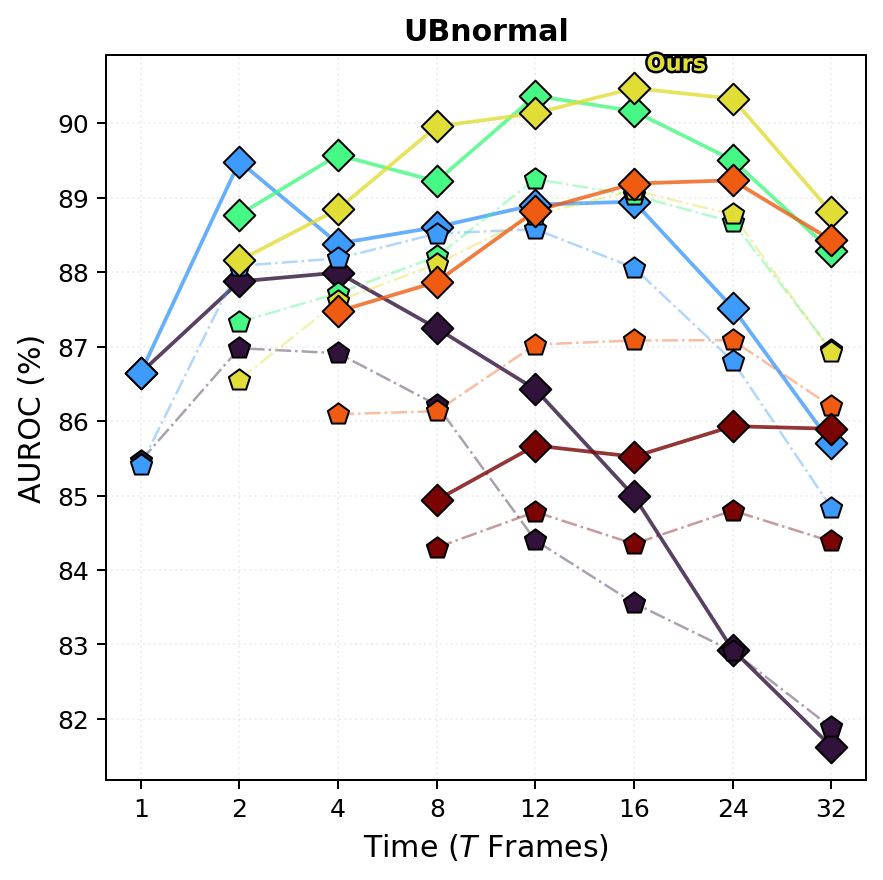}
  \end{minipage}\hfill
  \begin{minipage}[c]{0.4\textwidth}
    \includegraphics[width=\textwidth]{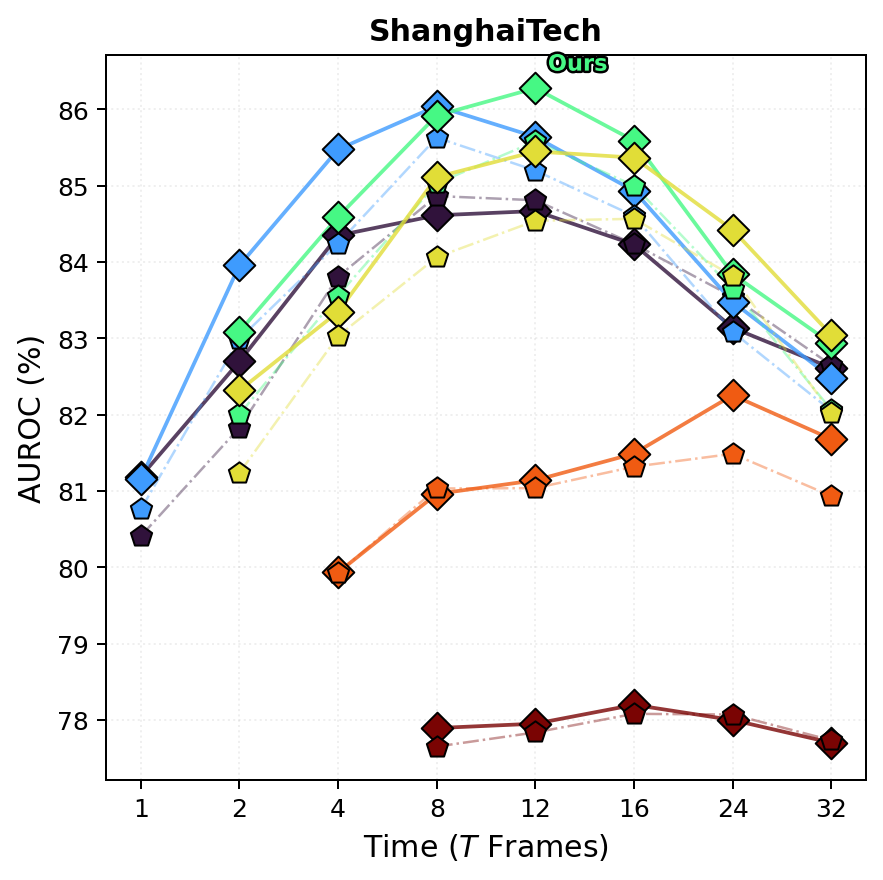}
  \end{minipage}\hfill
  \begin{minipage}[c]{0.20\textwidth}
    \includegraphics[scale=0.4]{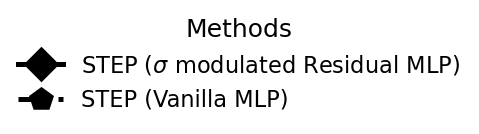}

    \vspace{5pt}

    \includegraphics[scale=0.4]{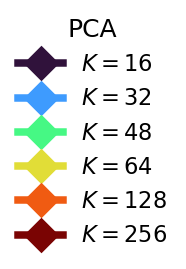}

  \end{minipage}

  \caption{\textbf{Architectural comparison.} We evaluate the performance differences across varying temporal lengths ($T$) and PCA dimensions ($K$) on \textbf{UBnormal} (left) and \textbf{ShanghaiTech} (right). The $\sigma$-modulated Residual MLP (diamonds) consistently outperforms the Vanilla MLP (pentagons) across almost all configurations, proving the benefit of explicitly conditioning the network on the noise scale.}
  \label{fig:arch_combined}
\end{figure}

In \cref{sec:architecture} of the main manuscript, we reported that integrating the noise scale $\sigma$ into the network via our $\sigma$-modulated Residual MLP yields a performance increase. \cref{fig:arch_combined} provides the exhaustive sweep confirming this architectural advantage. Across almost every $T$ and $K$ combination on both datasets, the $\sigma$-modulated architecture (diamonds) consistently outperforms the standard, unmodulated Vanilla MLP (pentagons) utilized in our STEP framework.

%% file: supp_03_evaluation_comprehensive.tex
\subsection{Extended State-of-the-Art Comparison}
\label{supp:sota_extended}

In \cref{sec:experiments} of the main manuscript, we primarily compared STEP against recent skeleton-based density estimation methods. In \cref{tab:sota_comparison_supp}, we provide an extended comparison that includes a wide variety of multi-modal and pixel-based methodologies. Methods utilizing deep features (\eg CLIP features, video features) (D) or motion (\eg modeled with optical flow) (M) often benefit from rich contextual background information. Remarkably, our purely skeleton-based approach (S) outperforms almost all multi-modal methods on both ShanghaiTech and UBnormal.

\textbf{PCA/kNN Baseline:} Furthermore, we include our own strong baseline, \textit{PCA/kNN}. This simple non-parametric approach, projecting temporal sequences onto the PCA space and scoring them based on the distance to their $k$-Nearest Neighbors ($k=1$) in the training set, achieves an impressive 81.3\% AUROC on UBnormal, outperforming complex prior state-of-the-art methods like SeeKer (77.9\%). However, our STEP framework, leveraging Denoising Score Matching (DSM), significantly elevates this performance to 90.1\%. We introduce a more detailed $T/K$ sweep later in \cref{sec:step_knn_pca} (\cref{fig:heatmap_st}, \cref{fig:heatmap_ub}).

\begin{table}[!h]
\setlength{\tabcolsep}{2pt}
\scriptsize
\centering
\caption{Extended state-of-the-art comparison using the AUROC (\%) metric. Input modality acronyms denote deep features (D), motion (M), and skeletons (S). STEP strictly outperforms all previous methods, including heavy multi-modal architectures, on both benchmarks. $^\dagger$reproduced by us.}
\label{tab:sota_comparison_supp}
  \begin{tabular}{ccc|lllll}
    \toprule
    \multicolumn{3}{c|}{Modality} & \multirow{2}{*}{Method} & \multicolumn{2}{l}{ShanghaiTech} & \multicolumn{2}{l}{UBnormal} \\
    D & M & S &  & Full & HR & Full & HR \\
    \midrule
    \multicolumn{8}{c}{\textbf{Multi-Modal / Pixel-Based Methods}} \\
    \midrule
    \checkmark & & & sRNN \cite{shanghaitech_luo2017revisit} & 68.0 & - & - & - \\
    \checkmark & & & Conv-AE \cite{hasan2016learning} & 70.4 & 69.8 & - & - \\
    \checkmark & & & LSA \cite{abati2019cvpr} & 72.5 & - & - & - \\
    \checkmark & \checkmark & & FFP \cite{liu2018future} & 72.8 & \underline{72.7} & - & - \\
    \checkmark & & & GCL \cite{zaheer2022generative} & 79.6 & - & - & - \\
    \checkmark & \checkmark & & CAE-SVM \cite{ionescu2017iccv} & 78.7 & - & - & - \\
    \checkmark & \checkmark & & VEC \cite{yu2020cloze} & 74.8 & - & - & - \\
    \checkmark & \checkmark & & $\text{HF}^2$ \cite{liu2021hybrid} & 76.2 & - & - & - \\
    \checkmark & \checkmark & & FPDM\cite{Yan_2023_ICCV} & 78.6 & - & \underline{62.7} & - \\
    \checkmark & & & SSMTL \cite{georgescu2021anomaly} & 82.7 & - & - & - \\
    \checkmark & \checkmark & & BA-AED \cite{georgescu2021background} & 82.7 & - & - & - \\
    \checkmark & & & SSMTL++ \cite{BARBALAU2023103656} & 83.8 & - & 62.1 & - \\
    \checkmark & & & Jigsaw \cite{wang2022video} & \underline{84.2} & \textbf{84.7} & 56.4 & - \\
    \checkmark & & & MULDE (Frame-centric) \cite{micorek24cvpr} & - & - & \textbf{72.8} & - \\
    \checkmark & \checkmark & \checkmark & MULDE (Object-centric) \cite{micorek24cvpr} & \textbf{86.7} & - & - & - \\
    \midrule
    \multicolumn{8}{c}{\textbf{Skeleton-Only Methods}} \\
    \midrule
    & & \checkmark & BiPOCO \cite{miracle2022arxiv} & - & 74.9 & 50.7 & 52.3 \\
    & & \checkmark & MPED-RNN \cite{morais2019cvpr} & 73.4 & 75.4 & 60.6 & 61.2 \\
    & & \checkmark & MTP \cite{rodrigues2020wacv} & 76.0 & 77.0 & - & - \\
    & & \checkmark & GEPC \cite{markovitz2020cvpr} & 76.1 & 74.8 & 53.4 & 55.2 \\
    & & \checkmark & PoseCVAE \cite{jain2020icpr} & - & 75.5 & - & - \\
    & & \checkmark & Normal Graph \cite{luo2020neurcomp} & - & 76.5 & - & - \\
    & & \checkmark & COSKAD \cite{flaborea2024pr} & - & 77.1 & 65.0 & 65.5 \\
    & & \checkmark & GCAE-LSTM \cite{li2021neurocmp} & - & 77.2 & - & - \\
    & & \checkmark & MoCoDAD \cite{Flaborea_2023_ICCV} & - & 77.6 & 68.3 & 68.4 \\
    & & \checkmark & MULDE (Skeleton, $T=1$) \cite{micorek24cvpr} & 78.5 & - & 80.6$^\dagger$ & - \\
    & & \checkmark & STG-NF \cite{Hirschorn_2023_ICCV} & \underline{85.9} & \underline{87.4} & 71.8 & 71.5 \\
    & & \checkmark & SeeKer \cite{delic2025seeker} & 85.5 & 86.9 & 77.9 & 78.9 \\
    \midrule
    & & \checkmark & PCA/kNN (Ours baseline) & 83.0 & - & \underline{81.3} & - \\
    \midrule
    & & \checkmark & \textbf{STEP (Ours)} & \textbf{86.2} $\pm$ 0.1 & \textbf{87.7} $\pm$ 0.1 & \textbf{90.1} $\pm$ 0.4 & \textbf{90.9} $\pm$ 0.4\\
    \bottomrule
\end{tabular}
\end{table}

\subsection{Additional AUC and AP Evaluation}
While AUROC is the standard metric for Video Anomaly Detection, it can sometimes mask false-positive sensitivities in highly imbalanced datasets. Therefore, we additionally report the Average Precision (AP) in \cref{tab:aucap}. Results are taken from SeeKer~\cite{delic2025seeker}. STEP consistently outperforms our two closest competitors on AP across both splits of both datasets. Notably, on the real-world ShanghaiTech dataset, we observe an even more pronounced margin of improvement in AP (+4.3\% over SeeKer on the Full set) than in AUROC.

\begin{table}[t]
\caption{Comparison on AUROC (\%) and Average Precision (AP\,\%) on UBnormal and ShanghaiTech. Best results are in \textbf{bold}, and second-best are \underline{underlined}.}
\label{tab:aucap}
\centering
\scriptsize
\setlength{\tabcolsep}{6pt}
\begin{tabular}{llcccc}
\toprule
\multirow{2}{*}{Dataset} & \multirow{2}{*}{Method} & \multicolumn{2}{c}{AUROC} & \multicolumn{2}{c}{AP} \\
\cmidrule(lr){3-4} \cmidrule(lr){5-6}
& & Full & HR & Full & HR \\
\midrule
\multirow{3}{*}{UBnormal}
& STG-NF \cite{Hirschorn_2023_ICCV} & 71.8 & 71.5 & 62.7 & 67.2 \\
& SeeKer \cite{delic2025seeker} & \underline{77.9} & \underline{78.9} & \underline{80.3} & \underline{79.8} \\
& \textbf{STEP} & \textbf{90.3} & \textbf{90.9} & \textbf{91.6} & \textbf{91.7} \\
\midrule
\multirow{3}{*}{ShanghaiTech}
& STG-NF \cite{Hirschorn_2023_ICCV} & \underline{85.9} & \underline{87.4} & 77.6 & 81.4 \\
& SeeKer \cite{delic2025seeker} & 85.5 & 86.9 & \underline{80.0} & \underline{81.5} \\
& \textbf{STEP } & \textbf{86.2} & \textbf{87.7} & \textbf{84.3} & \textbf{85.2} \\
\bottomrule
\end{tabular}
\end{table}

\subsection{STEP vs. kNN/PCA Baseline Evaluation}
\label{sec:step_knn_pca}
\begin{figure}[!htbp]
  \centering
  \includegraphics[width=\textwidth]{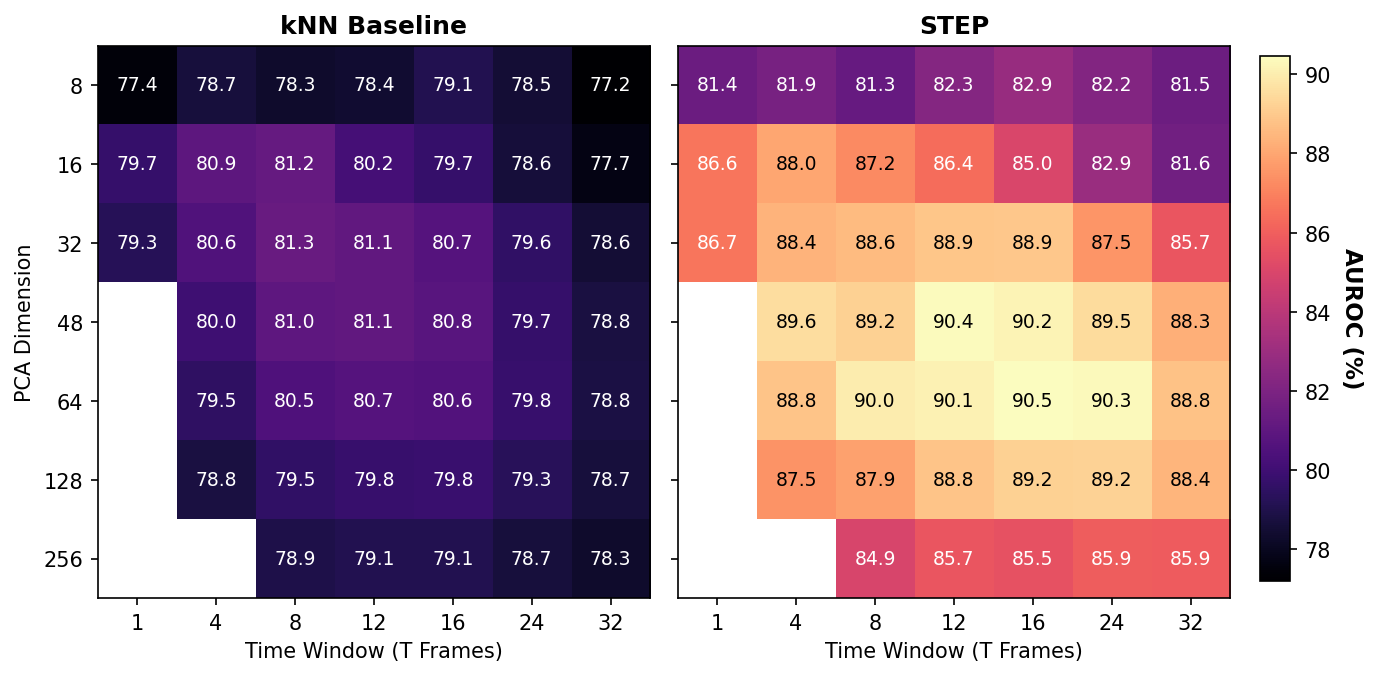}
  \caption{\textbf{UBnormal $T/K$ Heatmaps:} The best performing PCA/kNN configurations do not strictly align with our EBM's optimal configuration. However, even the simple PCA/kNN baseline comfortably outperforms the current SOTA.}
  \label{fig:heatmap_ub}
\end{figure}

\begin{figure}[thbp]
  \centering
  \includegraphics[width=\textwidth]{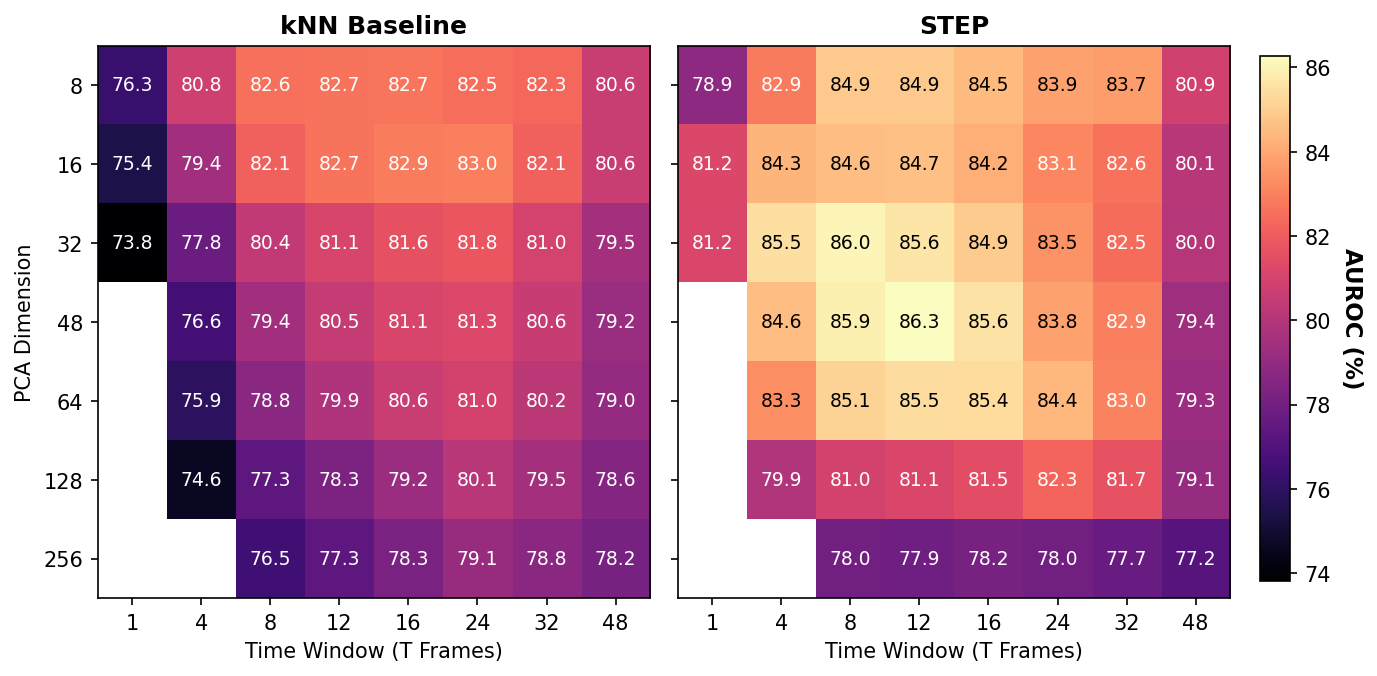}
  \caption{\textbf{ShanghaiTech $T/K$ Heatmaps:} Similar to UBnormal, the optimal temporal and dimensional capacities for distance-based scoring (kNN) differ from density-based scoring (STEP).}
  \label{fig:heatmap_st}
\end{figure}

To establish the complexity required for anomaly scoring, we extensively evaluated a non-parametric $k$-Nearest Neighbors (kNN) baseline operating directly on the extracted PC-space. As visualized in \cref{fig:heatmap_ub} and \cref{fig:heatmap_st}, this simple PCA/kNN ($k=1$) baseline is a surprisingly strong competitor, achieving performance that rivals or exceeds current skeleton-based state-of-the-art methods.

Interestingly, the heatmaps reveal that the optimal configuration grid ($T$, $K$) for the kNN approach does not directly align with the settings for STEP. While kNN generally favors slightly lower PCA dimensions to avoid distance metric degradation in higher dimensions, the Energy-Based Model inside STEP thrives with higher capacity ($K=32, 48, 64$), utilizing the extra dimensions to model nuanced motion boundaries. Because these optimal capacities decouple, we strictly evaluate and select our final $T/K$ configuration based on the separate UBnormal validation split.

\subsection{Impact of Minimum and Maximum Noise Scales ($\sigma_{low}, \sigma_{high}$)}
To complement the component breakdown \cref{tab:unified_ablation} of the main manuscript, we provide the full data tables for the $\sigma_{low}$ (\cref{tab:sigma_low_sweep}) and $\sigma_{high}$ (\cref{tab:sigma_high_sweep}) boundary ablations. The base settings are $\sigma_{low}=0.1, \sigma_{high}=1$.

As discussed, lowering the minimum noise scale too far ($\sigma_{low}=10^{-4}$) forces the network to model microscopic tracking jitter rather than macroscopic pose validity. Conversely, setting it too high ($\sigma_{low}=0.5$) over-smooths the density estimate, blurring the fine-grained boundaries of normal motion.

For the upper bound ($\sigma_{high}$), the tables reveal an interesting dataset-specific divergence that aligns perfectly with the nature of their respective anomalies. On ShanghaiTech, which primarily contains subtle kinematic deviations (\eg, riding a bicycle or pushing), a tighter maximum scale ($\sigma_{high}=0.2$) actually yields slightly better performance. This lower maximum bound focuses the network's capacity strictly around the immediate boundary of normal motion. However, on UBnormal, which features extreme synthetic pose violations (\eg, falling flat or drastic erratic movements), this narrow coverage is insufficient. A tight upper bound fails to provide gradient signals in the distant, sparse regions of the PC-space, leaving those extreme anomalies unscored. Ultimately, our universally selected bounds ($\sigma_{low}=0.1, \sigma_{high}=1.0$) provide an overall robust balance. They maintain a structurally sound energy landscape capable of detecting both subtle and extreme anomalies across diverse benchmarks without requiring dataset-specific tuning.

\begin{table}[th]
\setlength{\tabcolsep}{9pt}
\scriptsize
\centering
\caption{Impact of minimum noise scale ($\sigma_{low}$) across different temporal windows ($T$) and PCA dimensions ($K$) with a fixed $\sigma_{high}=1$. Results are in AUROC (\%). For a given $T/K$ setting we mark the best performing in \textbf{bold}, \underline{underline} the second best and the third best is in \textit{italic}.}
\label{tab:sigma_low_sweep}
\begin{tabular}{cc ccc  ccc}
\toprule
\multirow{2}{*}{$T$} & $\sigma_{high}=1$ & \multicolumn{3}{c}{UBnormal ($K$)} & \multicolumn{3}{c}{ShanghaiTech ($K$)} \\
\cmidrule(lr){2-2} \cmidrule(lr){3-5} \cmidrule(lr){6-8}
 & $\sigma_{low}$ & 32 & 48 & 64 & 32 & 48 & 64 \\
\midrule
\multirow{5}{*}{12}
 & 1e-4  & \underline{89.13} & \textit{89.56} & 89.38 & \underline{85.52} & \underline{86.05} & 85.22 \\
 & 0.001 & \textbf{89.33} & 89.24 & \textit{89.48} & \textit{85.43} & \textit{85.97} & \textit{85.42} \\
 & 0.01  & \textit{88.93} & \underline{89.70} & \underline{89.48} & 84.98 & 85.85 & \textbf{85.68} \\
 & 0.1   & 88.91 & \textbf{90.37} & \textbf{90.54} & \textbf{85.64} & \textbf{86.28} & \underline{85.45} \\
 & 0.5   & 86.37 & 86.91 & 86.47 & 83.72 & 83.88 & 82.65 \\
\midrule
\multirow{5}{*}{16}
 & 1e-4  & 88.45 & \textit{89.57} & \textit{89.39} & \textit{84.43} & \underline{85.44} & 84.90 \\
 & 0.001 & \textit{88.60} & 89.19 & 88.98 & \underline{84.46} & 85.31 & \textit{85.25} \\
 & 0.01  & \textbf{89.17} & \underline{89.58} & \underline{89.80} & 84.35 & \textit{85.41} & \textbf{85.64} \\
 & 0.1   & \underline{88.95} & \textbf{89.67} & \textbf{90.60} & \textbf{84.93} & \textbf{85.59} & \underline{85.37} \\
 & 0.5   & 85.57 & 86.76 & 86.34 & 83.35 & 83.66 & 82.86 \\
\midrule
\multirow{5}{*}{24}
 & 1e-4  & \textit{88.60} & 88.69 & \textit{88.91} & \textit{82.78} & \textit{83.55} & 83.80 \\
 & 0.001 & \underline{88.63} & \underline{89.05} & 88.72 & \underline{83.26} & \textbf{83.98} & \textit{84.11} \\
 & 0.01  & \textbf{89.26} & \textit{88.98} & \underline{89.44} & 82.52 & 83.54 & \underline{84.22} \\
 & 0.1   & 87.52 & \textbf{89.51} & \textbf{90.33} & \textbf{83.48} & \underline{83.84} & \textbf{84.42} \\
 & 0.5   & 84.18 & 85.50 & 86.27 & 82.78 & 82.79 & 82.43 \\
\bottomrule
\end{tabular}
\end{table}

\begin{table}[th]
\setlength{\tabcolsep}{9pt}
\scriptsize
\centering
\caption{Impact of maximum noise scale ($\sigma_{high}$) across different temporal windows ($T$) and PCA dimensions ($K$) with a fixed $\sigma_{low}=0.1$. Results are in AUROC (\%). For a given $T/K$ setting we mark the best performing in \textbf{bold}, \underline{underline} the second best and the third best is in \textit{italic}.}
\label{tab:sigma_high_sweep}
\begin{tabular}{cc ccc  ccc}
\toprule
\multirow{2}{*}{$T$} & $\sigma_{low}=0.1$ & \multicolumn{3}{c}{UBnormal ($K$)} & \multicolumn{3}{c}{ShanghaiTech ($K$)} \\
\cmidrule(lr){2-2} \cmidrule(lr){3-5} \cmidrule(lr){6-8}
 & $\sigma_{high}$ & 32 & 48 & 64 & 32 & 48 & 64 \\
\midrule
\multirow{4}{*}{12}
 & 0.2 & \textbf{89.49} & 89.87 & 90.14                   & \textbf{86.03} & \underline{86.38} & \underline{85.77} \\
 & 0.5 & \underline{89.48} & \textbf{90.57} & \textit{90.24} & \textit{85.74} & \textbf{86.49} & \textbf{86.02} \\
 & 1   & 88.91 & \underline{90.37} & \textbf{90.54}       & 85.64 & \textit{86.28} & \textit{85.54} \\
 & 2   & \textit{89.12} & \textit{89.89} & \underline{90.43} & \underline{85.77} & 86.10 & 85.29 \\
\midrule
\multirow{4}{*}{16}
 & 0.2 & \textit{88.50} & 89.00 & 89.71                   & \textbf{85.08} & \textbf{86.05} & \textbf{85.96} \\
 & 0.5 & \textbf{89.49} & \textit{89.36} & \textit{90.41} & \textit{84.84} & \underline{85.66} & \underline{85.75} \\
 & 1   & \underline{88.95} & \underline{89.67} & \textbf{90.60} & \underline{84.93} & \textit{85.59} & \textit{85.37} \\
 & 2   & 88.27 & \textbf{89.70} & \underline{90.51}       & 84.77 & 85.55 & 84.84 \\
\midrule
\multirow{4}{*}{24}
 & 0.2 & \underline{87.42} & \textit{89.20} & 88.17       & \textit{83.83} & \textbf{84.57} & \textbf{84.76} \\
 & 0.5 & \textit{87.26} & 89.04 & \textit{89.08}          & \underline{83.85} & \underline{84.27} & \textit{84.36} \\
 & 1   & 86.96 & \underline{89.51} & \textbf{90.33}       & 83.48 & 83.84 & \underline{84.42} \\
 & 2   & \textbf{87.87} & \textbf{89.74} & \underline{89.78} & \textbf{83.88} & \textit{84.08} & 84.11 \\
\bottomrule
\end{tabular}
\end{table}

\subsection{Pose Confidence Filtering Before Training}
In \cref{sec:confidence} of the main manuscript, we advocated for a soft, sequence-level confidence weighting loss over a hard pre-training filter. \cref{tab:drop_pose_conf} justifies this choice.

A clear trend emerges based on the PCA capacity ($K$). For lower capacities (e.g., $K=16$), the PCA bottleneck is extremely tight. Feeding severely noisy or broken poses into this limited representation corrupts it; therefore, applying a moderate confidence filter (e.g., dropping poses with conf $<0.2$) often \textit{improves} performance at $K=16$.

However, at our typical operating range at larger capacities ($K=32, 48, 64$), the PC-space has the dimensional bandwidth to properly encode complex, high-variance motion. Aggressively dropping poses (\eg, $<0.4$) discards a massive amount of valid, albeit fast or partially occluded, normal human motion. Depriving the network of this data creates blind spots in the high-dimensional density basins, leading to a catastrophic performance collapse (\eg dropping from 90.54\% to 86.10\% on UBnormal at $T=12, K=64$). Our soft sequence-level weighting entirely circumvents this trade-off, utilizing the full breadth of the training data while naturally suppressing the gradient influence of severe tracking failures.

\begin{table}[th]
\setlength{\tabcolsep}{5pt}
\scriptsize
\centering
\caption{Impact of dropping noisy poses during training across different temporal windows ($T$) and PCA dimensions ($K$). Poses below the confidence threshold are discarded. Results are in AUROC (\%). For a given $T/K$ setting we mark the best performing in \textbf{bold}, \underline{underline} the second best and the third best is in \textit{italic}.}
\label{tab:drop_pose_conf}
\begin{tabular}{cc cccc cccc}
\toprule
\multirow{2}{*}{$T$} & \multirow{2}{*}{Drop Conf.} & \multicolumn{4}{c}{UBnormal ($K$)} & \multicolumn{4}{c}{ShanghaiTech ($K$)} \\
\cmidrule(lr){3-6} \cmidrule(lr){7-10}
 & & 16 & 32 & 48 & 64 & 16 & 32 & 48 & 64 \\
\midrule
\multirow{4}{*}{8}
 & 0   & \textit{87.25} & \textit{88.61} & \underline{89.22} & \underline{89.96} & \textit{84.61} & \underline{86.04} & \underline{85.92} & \underline{85.11} \\
 & 0.1 & \underline{87.25} & \underline{88.89} & \textit{89.14} & \textbf{90.27} & \underline{84.84} & \textbf{86.16} & \textbf{86.03} & \textbf{85.25} \\
 & 0.2 & \textbf{87.86} & \textbf{89.17} & \textbf{89.49} & \textit{89.23} & \textbf{84.98} & \textit{85.74} & \textit{85.81} & \textit{85.07} \\
 & 0.4 & 86.37 & 86.43 & 86.51 & 85.43 & 84.12 & 84.95 & 84.43 & 83.36 \\
\midrule
\multirow{4}{*}{12}
 & 0   & \underline{86.43} & \textit{88.91} & \underline{90.37} & \textbf{90.54} & \underline{84.67} & \textit{85.64} & \underline{86.28} & \textbf{85.45} \\
 & 0.1 & \textit{86.28} & \underline{89.03} & \textbf{90.49} & \underline{90.45} & \textbf{84.80} & \underline{85.71} & \textbf{86.35} & \underline{85.42} \\
 & 0.2 & \textbf{86.72} & \textbf{89.78} & \textit{89.60} & \textit{90.03} & \textit{84.62} & \textbf{85.87} & \textit{86.00} & \textit{85.41} \\
 & 0.4 & 85.13 & 85.92 & 85.95 & 86.10 & 84.04 & 85.03 & 84.30 & 83.59 \\
\midrule
\multirow{4}{*}{16}
 & 0   & \underline{84.99} & \textbf{88.95} & \underline{89.67} & \textbf{90.60} & \underline{84.23} & \underline{84.93} & \underline{85.59} & \textbf{85.37} \\
 & 0.1 & 84.08 & \textit{88.63} & \textbf{90.21} & \underline{90.34} & \textbf{84.42} & \textit{84.85} & \textbf{85.68} & \underline{85.35} \\
 & 0.2 & \textbf{85.70} & \underline{88.79} & \textit{89.60} & \textit{90.23} & \textit{83.95} & \textbf{85.07} & \textit{85.48} & \textit{85.02} \\
 & 0.4 & \textit{84.46} & 84.71 & 85.42 & 86.06 & 83.57 & 84.57 & 84.14 & 83.39 \\
\midrule
\multirow{4}{*}{24}
 & 0   & \textit{82.93} & \underline{87.52} & \underline{89.51} & \textbf{90.33} & \underline{83.14} & \textit{83.48} & \textit{83.84} & \textbf{84.42} \\
 & 0.1 & 82.91 & \textit{87.30} & \textit{89.24} & \underline{90.26} & \textbf{83.42} & \underline{83.55} & \underline{84.17} & \underline{84.10} \\
 & 0.2 & \textbf{85.24} & \textbf{87.68} & \textbf{90.09} & \textit{90.17} & \textit{82.95} & \textbf{84.19} & \textbf{84.29} & \textit{84.02} \\
 & 0.4 & \underline{84.30} & 84.06 & 85.55 & 85.52 & 82.72 & 83.14 & 83.15 & 82.63 \\
\bottomrule
\end{tabular}
\end{table}

\subsection{Inference Protocols and Aggregation}
In \cref{tab:aggregation_comprehensive}, we present an evaluation of our framework across different inference protocols to highlight its stability. First, we contrast \textbf{Online (On)} and \textbf{Backtrack (BT)} inference. While both protocols require observing a full temporal window of $T$ frames to compute a score, they differ fundamentally in how that score is assigned. The Backtrack protocol introduced by SeeKer~\cite{delic2025seeker} is an offline, non-causal setting that evaluates the segment and retroactively applies the computed score to all past frames within that window, merging scores where necessary. This inherently introduces a temporal lag of $T$ frames, as the anomaly score for a given frame is only finalized once the entire future window has been observed. Conversely, Online inference is causal: it evaluates the preceding $T$ frames but immediately assigns the resulting score to the current frame $t$ or the center ($T/2$) of the temporal window.

Furthermore, we evaluate the impact of temporal 1D-Gaussian smoothing. While smoothing improves overall metrics, as is standard practice in skeleton-based VAD literature (\eg, STG-NF, SeeKer, MULDE), it relies on a smoothing window that introduces an additional time shift. Consequently, smoothed anomaly scores can only be reported with a (short) delay. Because STEP operates at real-time speeds, the \textit{Online Raw} metric is the only true zero-delay protocol representative of a live, real-world deployment scenario. Strikingly, even under this strict causal and unsmoothed constraint, our raw scores remain highly competitive.

Furthermore, \cref{tab:agg_ablation} and \cref{tab:aggregation_comprehensive} explicitly evaluate our \textit{Agg-Max} strategy (\cref{eq:aggregation} of the main manuscript) against alternative multiscale heuristics: \textit{Agg-Sum} (replacing the maximum operation over noise scales $i$ with a summation) and \textit{Best Individual} (retrospectively selecting the single highest-performing noise scale $\sigma$). Conceptually, anomalies manifest at vastly different structural frequencies. In a real-world deployment, we do not know \textit{a priori} which noise scale will optimally capture a given anomaly. Applying an \textit{Agg-Sum} strategy risks severe signal dilution; a strong anomaly peak at one specific scale is mathematically washed out by low, uninformative energies across the remaining scales (dropping to 87.2\% at $K=48$). \textit{Agg-Max} bypasses these issues by acting as a continuous logical OR gate, flagging the sequence if it exhibits anomalous behavior at \textit{any} structural frequency. Strikingly, this parameter-free approach matches or slightly exceeds the retrospectively chosen best individual scale, yielding a highly robust inference pipeline.

\begin{table}[!htbp]
\setlength{\tabcolsep}{4pt}
\scriptsize
\centering
\caption{Impact of the multiscale aggregation strategy on the UBnormal dataset ($T=12$). \textit{Agg-Max} successfully captures anomalies across all structural frequencies without diluting the energy signal, drastically outperforming summation (\textit{Agg-Sum}) and competing directly with the retrospectively chosen best individual scale.}
\label{tab:agg_ablation}
\begin{tabular}{@{}l | c c c c c c c@{}}
\toprule
\multirow{2}{*}{\textbf{Aggregation Strategy}} & \multicolumn{7}{c}{\textbf{PCA Manifold Dimension ($K$)}} \\
 & 8 & 16 & 32 & \textbf{48} & 64 & 128 & 256 \\ \midrule
Agg-Sum & 79.7 & 83.3 & 87.2 & 87.2 & 87.1 & 84.8 & 81.5 \\
Best Individual $\sigma$ & 81.9 & \textbf{86.9} & \textbf{89.6} & 90.3 & \textbf{90.1} & \textbf{89.2} & 85.3 \\ \midrule
\textbf{Agg-Max (Ours)} & \textbf{82.3} & 86.4 & 88.9 & \textbf{90.4} & \textbf{90.1} & 88.8 & \textbf{85.7} \\ \bottomrule
\end{tabular}
\end{table}

\begin{table}[!htbp]
\centering
\caption{Comprehensive evaluation of aggregation strategies across different inference protocols on the Full and Human-Related (HR) splits. We evaluate both offline/non-causal Backtrack (BT) and real-time/causal Online (On) settings, with and without temporal 1D-Gaussian smoothing. Results are in AUROC (\%) and AP (\%). We mark the best performing strategy in \textbf{bold}, \underline{underline} the second best, and the third best is in \textit{italic}.}
\label{tab:aggregation_comprehensive}
\resizebox{\textwidth}{!}{
\begin{tabular}{l | cc cc cc cc | cc cc cc cc}
\toprule
\multirow{3}{*}{\textbf{Aggregation}} & \multicolumn{8}{c|}{\textbf{UBnormal}} & \multicolumn{8}{c}{\textbf{ShanghaiTech}} \\
\cmidrule(lr){2-9} \cmidrule(l){10-17}
 & \multicolumn{2}{c}{BT Raw} & \multicolumn{2}{c}{BT Smooth} & \multicolumn{2}{c}{On Raw} & \multicolumn{2}{c|}{On Smooth} & \multicolumn{2}{c}{BT Raw} & \multicolumn{2}{c}{BT Smooth} & \multicolumn{2}{c}{On Raw} & \multicolumn{2}{c}{On Smooth} \\
\cmidrule(lr){2-3} \cmidrule(lr){4-5} \cmidrule(lr){6-7} \cmidrule(lr){8-9} \cmidrule(lr){10-11} \cmidrule(lr){12-13} \cmidrule(lr){14-15} \cmidrule(l){16-17}
 & AUC & AP & AUC & AP & AUC & AP & AUC & AP & AUC & AP & AUC & AP & AUC & AP & AUC & AP \\
\midrule
\multicolumn{17}{c}{\textbf{Full Test-set}} \\
\midrule
\textit{Agg-Max}             & \underline{87.73} & \underline{91.31} & \textbf{90.99} & \textbf{92.28} & \underline{85.03} & \textbf{89.84} & \textbf{90.59} & \textbf{91.90} & \underline{83.93} & \underline{81.08} & \underline{85.39} & \underline{83.20} & \underline{83.55} & \underline{81.41} & \textbf{86.32} & \underline{84.52} \\
\textit{Agg-Sum}             & 85.76 & 89.58 & 88.17 & 91.41 & 83.53 & 88.45 & 86.92 & 90.64 & 81.95 & 79.08 & 83.71 & 81.36 & 81.86 & 79.48 & 84.88 & 82.75 \\
\textit{Agg-Mean-Top2}      & \textit{87.67} & \textit{91.27} & \underline{90.89} & \underline{92.21} & \textit{84.98} & \textit{89.80} & \underline{90.50} & \underline{91.84} & \textit{83.78} & \textit{80.95} & \textit{85.26} & \textit{83.07} & \textit{83.41} & \textit{81.27} & \textit{86.20} & \textit{84.38} \\
\textit{Agg-Mean-Top3}       & 87.56 & 91.19 & 90.72 & 92.10 & 84.89 & 89.74 & 90.36 & 91.74 & 83.64 & 80.85 & 85.14 & 82.97 & 83.29 & 81.15 & 86.10 & 84.26 \\
\textit{Best ind. $\sigma$}      & \textbf{87.84} & \textbf{91.37} & \textit{90.85} & \underline{92.21} & \textbf{85.10} & \textbf{89.84} & \textit{90.44} & \textit{91.81} & \textbf{84.05} & \textbf{82.03} & \textbf{85.51} & \textbf{83.82} & \textbf{83.61} & \textbf{81.91} & \underline{86.29} & \textbf{84.62} \\
\midrule
\multicolumn{17}{c}{\textbf{Human-Related Split}} \\
\midrule
\textit{Agg-Max}             & \underline{88.99} & \underline{91.71} & \textbf{91.81} & \textbf{92.53} & \underline{86.20} & \textbf{90.17} & \textbf{91.34} & \textbf{92.10} & \underline{85.64} & \underline{82.02} & \underline{87.27} & \underline{84.24} & \underline{85.12} & \underline{82.31} & \textbf{87.97} & \underline{85.45} \\
\textit{Agg-Sum}             & 87.07 & 89.95 & 89.55 & 91.89 & 84.76 & 88.77 & 88.27 & 91.09 & 83.63 & 80.06 & 85.58 & 82.48 & 83.43 & 80.44 & 86.61 & 83.76 \\
\textit{Agg-Mean-Top2}      & \textit{88.94} & \textit{91.67} & \textit{91.72} & \textit{92.47} & \textit{86.16} & \textit{90.14} & \underline{91.28} & \underline{92.06} & \textit{85.48} & \textit{81.89} & \textit{87.13} & \textit{84.12} & \textit{84.98} & \textit{82.17} & \textit{87.85} & \textit{85.32} \\
\textit{Agg-Mean-Top3}      & 88.83 & 91.60 & 91.57 & 92.36 & 86.09 & 90.08 & 91.15 & 91.98 & 85.33 & 81.80 & 87.01 & 84.03 & 84.85 & 82.06 & 87.74 & 85.20 \\
\textit{Best ind. $\sigma$}     & \textbf{89.08} & \textbf{91.75} & \underline{91.74} & \underline{92.50} & \textbf{86.27} & \textbf{90.17} & \textit{91.26} & \textit{92.03} & \textbf{85.74} & \textbf{83.08} & \textbf{87.40} & \textbf{84.98} & \textbf{85.17} & \textbf{82.92} & \underline{87.94} & \textbf{85.66} \\
\bottomrule
\end{tabular}
}
\end{table}

\subsection{Cross-Dataset Generalization}
\label{sec:cross_dataset}

To evaluate the robustness and generalization capabilities of the learned energy landscapes, we perform a cross-dataset evaluation. Specifically, we take the STEP model trained exclusively on the normal training split of the ShanghaiTech dataset and evaluate it directly on the UBnormal test set without any fine-tuning. We then perform the inverse experiment, evaluating the UBnormal-trained model on ShanghaiTech.

\begin{table}[!t]
\setlength{\tabcolsep}{5pt}
\footnotesize
\centering
\caption{\textbf{Cross-Dataset Evaluation:} We evaluate the generalization of STEP by testing models on datasets they were not trained on. Finally, we jointly train both datasets in a single model and evaluate on both datasets. Results reported are AUROC (\%).}
\label{tab:cross_dataset}
\begin{tabular}{l | c c c}
\toprule
\textbf{Test} \textbackslash \textbf{Train} & ShanghaiTech & UBnormal & Both \\
\midrule
ShanghaiTech & 86.3 & 77.2 & 86.0 \\
UBnormal     & 82.4 & 90.6 & 88.9 \\
\bottomrule
\end{tabular}
\end{table}

\textbf{Analysis of Results.}
As shown in \cref{tab:cross_dataset}, STEP demonstrates strong cross-dataset generalization, despite the inherent domain gap between the datasets.
Notably, the model trained on ShanghaiTech generalizes exceptionally well to UBnormal, achieving an 82.4\% AUROC. This indicates that the compact PCA manifold and the resulting energy landscape learned from diverse, real-world pedestrian motion (ShanghaiTech) to successfully capture universal kinematic priors that directly apply to the synthetic avatars in UBnormal.

Conversely, the model trained on UBnormal achieves a 77.2\% AUROC when evaluated on ShanghaiTech. This slight asymmetry in generalization is expected. Because UBnormal is a purely synthetic dataset generated from a specific set of programmed animations, its underlying distribution of normal motion is inherently narrower than the highly diverse, multi-camera, real-world behaviors found in ShanghaiTech. Consequently, the UBnormal-trained model encounters unseen but benign real-world motions in the ShanghaiTech test set, resulting in slightly higher false positive rates. Overall, these results validate that STEP learns a fundamental, structurally sound representation of human kinematics rather than merely overfitting to dataset-specific backgrounds or camera angles.

\subsection{Evaluation MSAD-HR Dataset}
\label{sec:msad_hr}

We evaluate STEP on the real-world MSAD dataset~\cite{msad2024} to validate generalization to a third, independently collected benchmark. MSAD comprises 720 videos across 14 scene types, with 11 anomaly categories split into seven \textbf{human-related (HR)} categories---Assault, Fighting, People Falling, Robbery, Shooting, Traffic Accident, and Vandalism---and four \textbf{non-human} categories: Explosion, Fire, Object Falling, and Water Incident. A skeleton-based detector carries no signal for non-human events: if no person is detected, no pose features are extracted, and no anomaly score is produced. Full MSAD-benchmark evaluation, shown in \cref{tab:msad_hr}, thus artificially suppresses the AUROC, as the model is inherently blind to anomalies lacking human subjects.

\textbf{Evaluation protocols.} MSAD provides two official evaluation protocols. We report under protocol~(i) (self-supervised), which evaluates on all 360 held-out test clips (120 normal $+$ 240 abnormal). Protocol~(ii) (weakly-supervised held-out) reserves half the abnormal clips for weakly-supervised training.

\textbf{Training setup.} We train on the 360 MSAD normal clips using poses extracted with the same AlphaPose~\cite{alphapose} pipeline as STG-NF~\cite{Hirschorn_2023_ICCV} and SeeKer~\cite{delic2025seeker}, ensuring a fair comparison. We follow the training procedure from ShanghaiTech described in \cref{sec:experiments} of the main manuscript: $T{=}12$, $K{=}48$, fixed 400-epoch cosine annealing without early stopping, and an initial learning rate of $2{\times}10^{-4}$.

\textbf{Results.} On the full MSAD benchmark (including non-HR anomalies), STEP achieves 59.6\% AUROC (Online Smooth), reflecting the systematic blindness of skeleton-based methods to non-human anomaly categories. Restricting to HR clips recovers this gap substantially: STEP achieves 74.1\% AUROC under UB-Style and 71.6\% under ST-Style (both Online Smooth), substantially improving over STG-NF (55.7\%) and SeeKer (61.1\%), whose HR split style is not specified.

\begin{table}[t]
\centering
\caption{
  AUROC (\%) on MSAD~\cite{msad2024} under evaluation protocol~(i): 360 held-out clips (120 normal $+$ 240 abnormal).
  \textbf{ST-Style} (as in ShanghaiTech): excludes non-HR clips. \textbf{UB-Style} (as in UBnormal): marks non-HR anomalous frames as normal.
  \textbf{BT}: Backtrack (non-causal). \textbf{On}: Online (causal, real-time).
  \textbf{Raw}/\textbf{Smooth}: without/with temporal 1D-Gaussian smoothing. $^\dagger$HR split style unspecified.
}
\small\setlength{\tabcolsep}{5pt}
\resizebox{\linewidth}{!}{
\begin{tabular}{l l cc | cc}
\toprule
\multirow{3}{*}{\textbf{Method}}
  & \multirow{3}{*}{\textbf{Variant}}
  & \multicolumn{4}{c}{\textbf{AUROC (\%)}} \\
  \cmidrule(lr){3-6}
  & & \multicolumn{2}{c|}{BT} & \multicolumn{2}{c}{On} \\
  \cmidrule(lr){3-4} \cmidrule(lr){5-6}
  & & Raw & Smooth & Raw & Smooth \\
\midrule
STG-NF~\cite{Hirschorn_2023_ICCV} & MSAD-HR & \multicolumn{4}{c}{55.7$^\dagger$} \\
SeeKer~\cite{delic2025seeker}      & MSAD-HR & \multicolumn{4}{c}{61.1$^\dagger$} \\
\midrule
\multirow{3}{*}{STEP (Ours)}
  & MSAD (includes all: HR and non-HR)                      & 57.6 & 59.9 & 56.4 & 59.6 \\
  & MSAD-HR / ST-Style (drop non-HR clips)             & 71.7 & 72.8 & 69.9 & 71.6 \\
  & MSAD-HR / UB-Style (relabeled non-HR clips to normal)           & 73.9 & 75.4 & 72.2 & 74.1 \\
\bottomrule
\end{tabular}
}
\label{tab:msad_hr}
\end{table}

\subsection{Runtime Efficiency}
\label{supp:runtime}

Since upstream pose extraction typically dominates the computational budget in skeleton-based Video Anomaly Detection pipelines, it is crucial that the downstream anomaly scoring mechanism adds minimal latency. We demonstrate that once poses are extracted and tracked, our proposed STEP framework introduces virtually negligible computational overhead.

Our lightweight architecture achieves this through a series of highly efficient, easily parallelizable operations: the flattened pose sequence is projected into the compact PC-space via a single matrix multiplication, whitened, and then passed through the Residual MLP in a single parallel batch across all $L=10$ noise scales. After standardizing the outputs per scale and applying our Agg-Max pooling per person, the final frame-level scores are complete. Crucially, because the Energy-Based Model operates strictly on the fixed-dimensional PCA bottleneck ($K$), the network's latency is completely decoupled from the temporal window size ($T$). As detailed in \cref{tab:efficiency}, inference speeds remain practically invariant whether the model evaluates 8 or 24 frames of temporal context, only the PCA projection is affected by the temporal window size.

For realistic, dense crowds of 50 persons per frame, STEP performs this entire scoring pipeline in under 1\,ms ($\sim$1026\,FPS), requiring a peak memory footprint of only 32.6\,MB on a single NVIDIA GTX 1080 GPU. Even in extreme, heavily crowded scenarios tracking up to 100 persons simultaneously, latency remains at just 1.62\,ms ($\sim$618\,FPS) with a minimal footprint of 44.9\,MB. These results confirm that STEP provides state-of-the-art anomaly detection at a fractional cost to the underlying pose extraction, making it highly viable for real-time, high-density monitoring on resource-constrained hardware.
\begin{table}[thbp]
\setlength{\tabcolsep}{8pt}
\scriptsize
\centering
\caption{Inference performance of STEP once poses are extracted. Latency covers the entire pipeline from raw coordinates to final frame-level scores ($L=10$ noise scales $\sigma$). $K$ denotes the PCA dimension and $T$ represents the temporal segment length.}
\label{tab:efficiency}
\begin{tabular}{cccccc}
\toprule
$K$ & $T$ & Persons/Frame & Latency (ms) & FPS & Peak Mem (MB) \\
\midrule
48 & 12 & 1   & 0.86 & 1167 & 21.5 \\
48 & 12 & 10  & 0.93 & 1077 & 22.8 \\
48 & 12 & 20  & 0.90 & 1109 & 25.2 \\
48 & 12 & 30  & 0.92 & 1087 & 28.6 \\
48 & 12 & 40  & 0.95 & 1051 & 30.0 \\
48 & 12 & 50  & 0.97 & 1026 & 32.6 \\
48 & 12 & 100 & 1.62 & 618 & 44.9 \\
48 & 12 & 500 & 7.45 & 134 & 143.2 \\
\midrule
48 & 8 & 20  & 0.91 & 1097 & 25.2 \\
48 & 12 & 20  & 0.93 & 1073 & 25.2 \\
48 & 16 & 20  & 0.94 & 1064 & 25.2 \\
48 & 24 & 20  & 0.92 & 1087 & 25.3 \\
\midrule
8 & 16 & 20  & 0.97 & 1027 & 24.9 \\
16 & 16 & 20  & 0.99 & 1005 & 25.0 \\
32 & 16 & 20  & 0.92 & 1090 & 25.1 \\
48 & 16 & 20  & 0.90 & 1111 & 25.2 \\
64 & 16 & 20  & 0.94 & 1062 & 25.3 \\
\bottomrule
\end{tabular}
\end{table}

%% file: supp_04_pca.tex
\section{PCA}
\label{supp:pca_insights}

In this section, we provide a deeper insight of the Principal Component Analysis (PCA), expanding upon the claims made in \cref{sec:pca,sec:confidence} of the main manuscript. We visualize latent traversals, demonstrate the PCA's implicit low-pass filtering properties, quantify its robustness to measurement noise, and explicitly compare it against non-linear Autoencoder alternatives.

\subsection{Latent Traversal on Static and Moving Poses}
\begin{figure}[thbp]
  \centering
  \includegraphics[width=\textwidth]{traversal_static_blue_with_ghost_v5_labeled.png}
  \includegraphics[width=\textwidth]{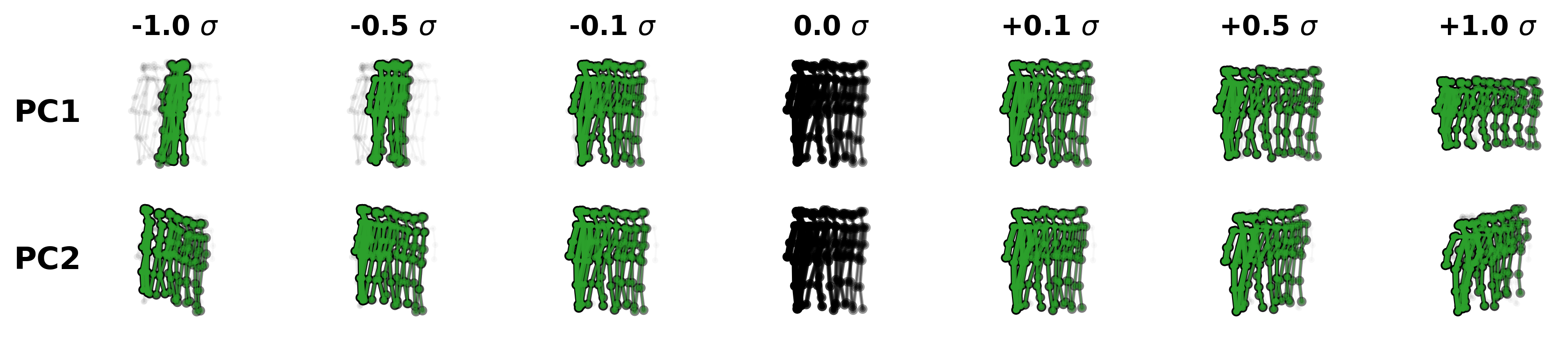}
  \caption{\textbf{Latent Traversal in the whitened PC-space.} We systematically add structured variance ($\sigma$) along the first two principal components. \textbf{Top (Static Pose):} Adding variance to a stationary person induces a coherent walking motion. \textbf{Bottom (Moving Pose):} Altering the latent coordinates of an already walking person smoothly alters the style and spatial orientation. The ghosting gray skeleton represents the zero-noise center pose for reference.}
  \label{fig:pca_traversal_static_moving}
\end{figure}

Our Denoising Score Matching (DSM) objective relies heavily on the assumption that adding isotropic Gaussian noise to the data representation yields physically plausible human poses. In \cref{fig:pca_traversal_static_moving}, we demonstrate the effect of systematically shifting the latent coordinates of a pose sequence along its first two principal components (PC1 and PC2).

Altering these components translates into continuous, semantically meaningful action variations. For a stationary person (top), adding variance physically manifests as inducing a coherent walking motion in opposite directions. For an already moving person (bottom), altering the latent coordinates smoothly alters the walking style and spatial orientation. In both cases, the kinematic structure and bone proportions remain structurally plausible without the severe distortions seen when noise is applied directly to raw joint coordinates. This structural preservation enables the Energy-Based Model to learn a stable, well-behaved density landscape.

\subsection{PCA as a Denoiser and Low-Pass Filter}
As described in the main text, projecting sequences through a PCA bottleneck acts as a denoiser. To visually and quantitatively validate this, we simulated tracking failures by injecting independent Gaussian noise ($\sigma_{pixel}$) directly into the raw pixel coordinates of a moving person sequence.

\begin{figure}[t]
  \centering
  \includegraphics[width=\textwidth]{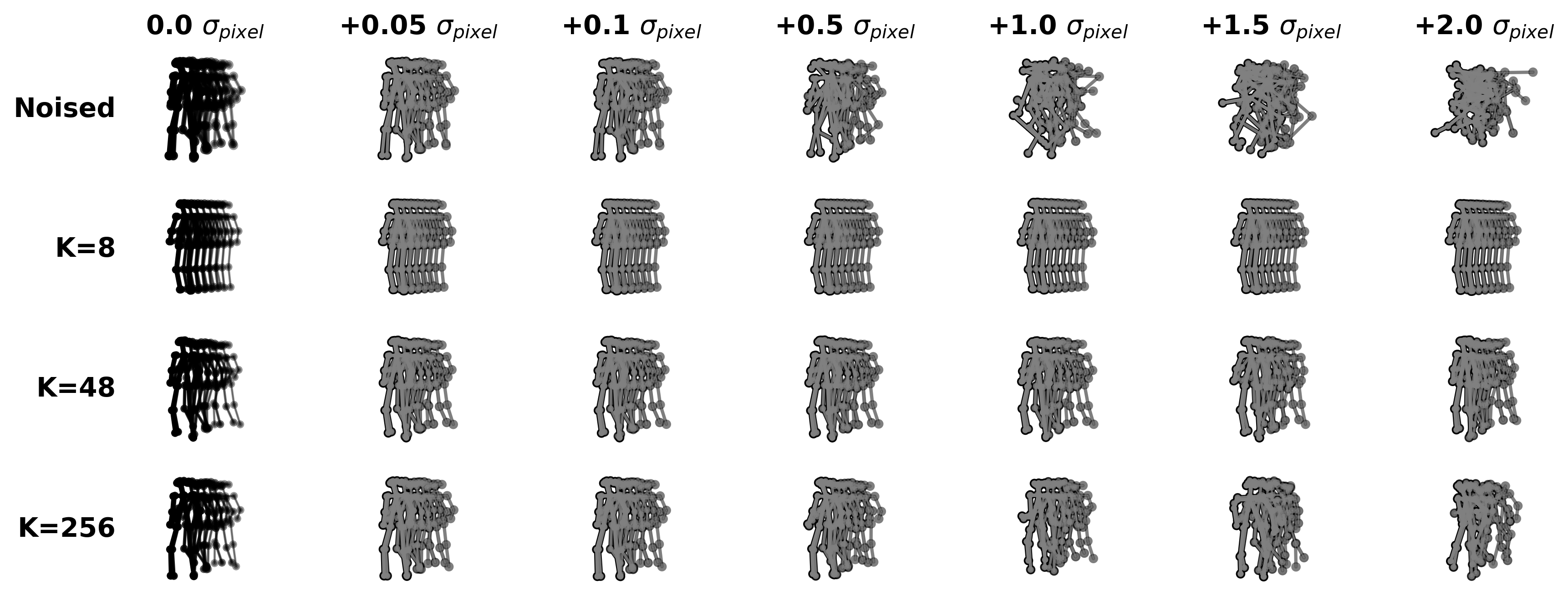}
  \caption{\textbf{PCA as a Low-Pass Filter.} \textbf{Top Row:} Adding independent Gaussian noise directly to raw pixel coordinates destroys the structure of a moving person. \textbf{Bottom Rows:} Projecting these corrupted sequences through various PCA bottlenecks filters the noise. An overly tight bottleneck ($K=8$) fails to recover complex motion, while a loose bottleneck ($K=256$) retains too many high-frequency components and remains sensitive to the noise. An appropriate $K=48$ configuration successfully removes unstructured jitter while preserving underlying motion.}
  \label{fig:pca_denoise_reconstruct}
\end{figure}

As shown in \cref{fig:pca_denoise_reconstruct}, adding this noise severely corrupts the structural integrity of the raw sequence. However, when we project this corrupted sequence into our PC-space and subsequently reconstruct it, the PCA successfully filters the noise. An overly tight bottleneck (\eg, $K=8$) over-smooths the data, failing to recover the complex walking motion. Conversely, a loose bottleneck (\eg, $K=256$) retains too many high-frequency components, passing the jitter directly to the Energy-Based Model. Our configuration of $K=48$ cleanly removes the unstructured measurement jitter while fully preserving the true, semantically meaningful human motion.

\begin{figure}[t]
  \centering
  \includegraphics[width=\textwidth]{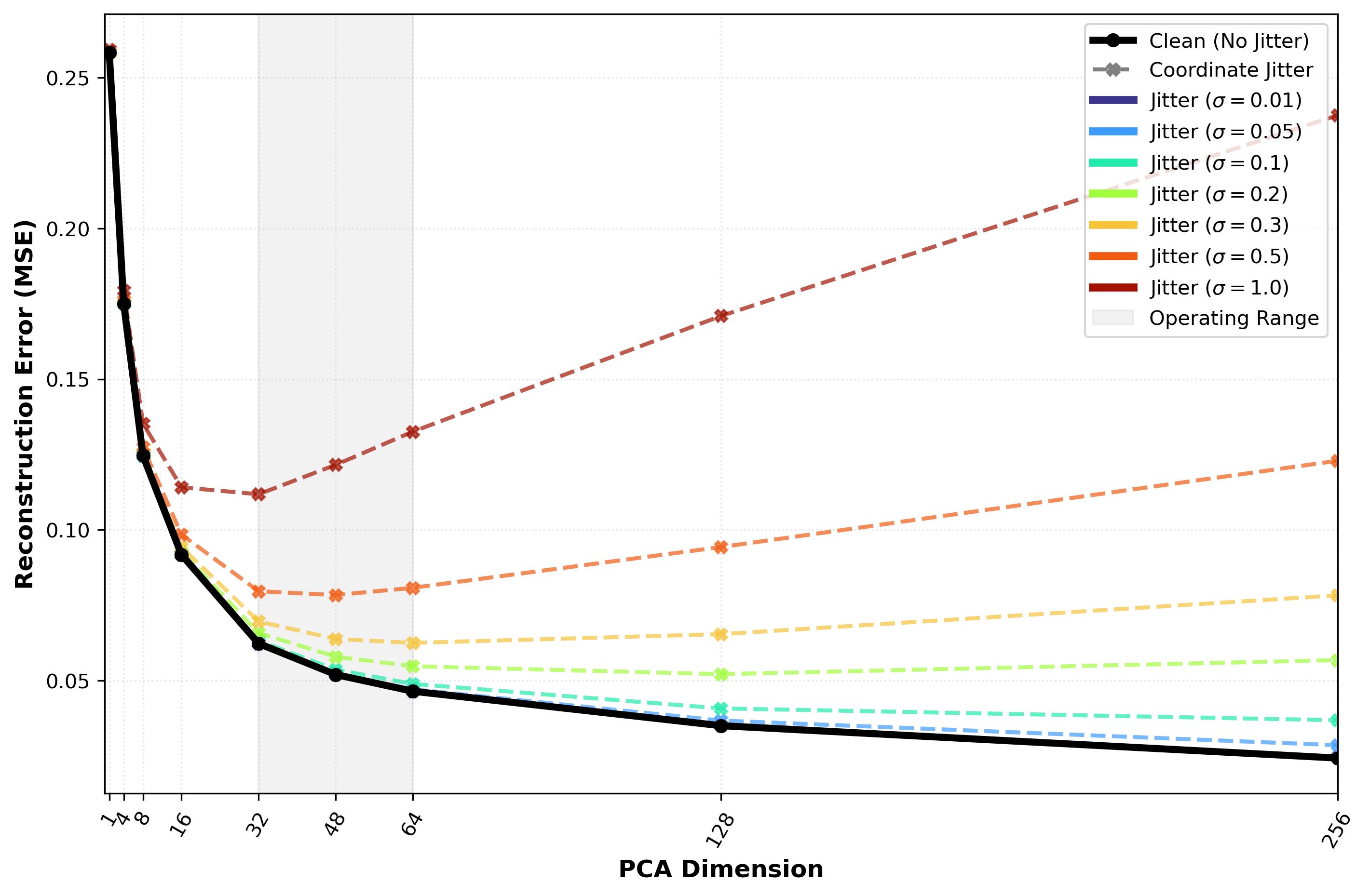}
  \caption{\textbf{Reconstruction Error under Input Noise ($T=12$, UBnormal).} As we increase the noise on the input pose coordinates, the poses can still be well reconstructed within the appropriate PCA capacity range of $K \in [32, 64]$. For larger PCA dimensions, the PC-space becomes too permissive and attempts to reconstruct the structural noise, causing the error to rise.}
  \label{fig:pca_reconstruct_denoise_across_pca}
\end{figure}

This behavior is quantitatively supported by the reconstruction error plot in \cref{fig:pca_reconstruct_denoise_across_pca}. As noise is injected into the raw input coordinates (dashed lines), sequences can still be well reconstructed up to a PCA dimension of $K \in [32, 64]$. Beyond this capacity, the projection manifold becomes too permissive and begins to reconstruct the high-frequency noise itself, causing the Mean Squared Error (MSE) on the reconstructed sequences to diverge.

\subsection{Quantitative Robustness to Pose Estimator Jitter}
\label{supp:jitter_robustness}
To explicitly quantify STEP's robustness to the inevitable tracking noise introduced by real-world pose extractors, we conducted a synthetic jitter ablation. During inference, we injected increasing magnitudes of independent Gaussian noise (standard deviation $\sigma_{pixel}$) directly into the raw spatial coordinates prior to the PCA projection. We evaluate the performance across our aggregation strategies in \cref{tab:jitter_robustness}.

The results highlight two critical mechanical behaviors of our framework:

\noindent \textbf{1. The Manifold as a Low-Pass Filter:} Because the spatial coordinates are normalized to the sequence bounding box scale, $\sigma_{pixel} = 0.05$ represents a subtle but realistic level of keypoint jitter, typical of what pose estimators naturally introduce under standard conditions (\eg, slight motion blur or minor occlusions). Despite this persistent underlying noise, the \textit{Agg-Max} AUROC remains stable (dropping by merely $0.2\%$ on UBnormal and $0.03\%$ on ShanghaiTech). This quantitatively confirms our claim that the compact PCA manifold successfully acts as a geometric low-pass filter, safely absorbing standard measurement jitter before it ever reaches the Energy-Based Model.

\noindent \textbf{2. Aggregation Behavior under Severe Degradation:} As the injected noise reaches more severe, unnatural levels ($\sigma_{pixel} \ge 0.2$), the \textit{Agg-Max} strategy predictably diverges and degrades faster than \textit{Agg-Sum} or the Best Individual scale. This elegantly illustrates the mechanics of multiscale score matching: severe, artificial high-frequency coordinate noise completely overwhelms the lowest energy scale ($\sigma_{low}$), causing it to output massive anomaly scores for every frame, regardless of whether the underlying motion is normal or abnormal. Because \textit{Agg-Max} acts as a logical OR gate, this floods the system with false positives. Conversely, \textit{Agg-Sum} dilutes these localized false positives across the sequence. Meanwhile, the \textit{Best Individual $\sigma$}, which is selected beforehand on the clean validation split, degrades more gracefully simply because this optimal validation scale typically corresponds to a mid-to-large $\sigma$ scale. These larger scales naturally focus on macroscopic structural anomalies and are inherently blind to microscopic, high-frequency coordinate jitter. Because \textit{Agg-Max} avoids the need for validation set calibration altogether, it remains robust within realistic operating bounds ($\sigma_{pixel} \le 0.1$) and yields the highest reliable, parameter-free performance.

\begin{table}[th]
\setlength{\tabcolsep}{4pt}
\scriptsize
\centering
\caption{Robustness to raw coordinate jitter during inference on UBnormal and ShanghaiTech ($T=12, K=48$). We inject independent Gaussian noise ($\sigma_{pixel}$) into the raw input coordinates prior to the PCA projection. Up to $\sigma_{pixel} = 0.1$, the PCA effectively acts as a low-pass filter, maintaining state-of-the-art performance. Tested on a single model with a fixed random seed.}
\label{tab:jitter_robustness}
\begin{tabular}{@{}l | c c c c c c c c@{}}
\toprule
\multirow{2}{*}{\textbf{Method}} & \multicolumn{8}{c}{\textbf{Pose Jitter} ($\sigma_{pixel}$)} \\ \cmidrule(l){2-9}
 & 0 & 0.01 & 0.05 & 0.10 & 0.20 & 0.30 & 0.50 & 1.00 \\ \midrule
\multicolumn{9}{@{}c}{\textbf{UBnormal}} \\ \midrule
\textit{Agg-Max} & 90.62 & 90.63 & 90.44 & 89.12 & 82.46 & 74.80 & 68.95 & 64.24 \\
\textit{Agg-Sum} & 87.57 & 87.58 & 87.70 & 87.67 & 86.41 & 82.14 & 72.21 & 62.89 \\
\textit{Best Ind.} & 90.49 & 90.49 & 90.21 & 89.53 & 86.98 & 85.50 & 79.13 & 65.82 \\ \midrule
\multicolumn{9}{@{}c}{\textbf{ShanghaiTech}} \\ \midrule
\textit{Agg-Max} & 86.32 & 86.33 & 86.29 & 86.00 & 80.36 & 75.31 & 71.90 & 68.29 \\
\textit{Agg-Sum} & 84.88 & 84.88 & 84.95 & 85.15 & 84.21 & 80.19 & 73.47 & 66.55 \\
\textit{Best Ind.} & 86.29 & 86.29 & 86.20 & 85.87 & 84.90 & 83.79 & 80.04 & 69.35 \\ \bottomrule
\end{tabular}
\end{table}

\subsection{PCA vs. Autoencoders (Linear vs. Non-Linear Projection)}
\label{supp:ae_ablation}

To evaluate the impact of the chosen projection, we compare our linear PCA space against non-linear projections. Specifically, we replaced the PCA projection with two deep non-linear baselines: a standard Autoencoder (AE) and a Variational Autoencoder (VAE). Both networks utilized a multi-layer architecture (Pose Input $\rightarrow$ Linear $\rightarrow$ BatchNorm $\rightarrow$ ReLU $\rightarrow$ Linear $\rightarrow$ Latent, reverse order for the decoder) and were trained for 200 epochs on the normal training set to match our PCA bottleneck dimension ($K=48$ at $T=12$). We subsequently trained our STEP framework on these extracted non-linear latents.

For a fixed random seed on the UBnormal benchmark, the linear PCA projection yields an AUROC of 90.5\% (\cref{tab:ae_vs_pca}). In contrast, replacing the linear projection with non-linear extractors degrades performance: a standard Autoencoder (AE) achieves 89.5\%, and a Variational Autoencoder (VAE) drops significantly to 84.9\%.

These initial extractors were trained under the standard One-Class Classification (OCC) setting, meaning they were optimized exclusively on normal training data. We hypothesize that the performance drop stems from a critical vulnerability of non-linear projections in this regime: Autoencoders trained only on normality risk mapping unseen anomalous sequences into regions of the normal latent space. This non-linear mode collapse effectively hides the anomalies, making it difficult for the subsequent Energy-Based Model (EBM) to distinguish and score them accurately.

To investigate this hypothesis, we leveraged the abnormal training sequences in the UBnormal dataset. We pretrained both the AE and VAE using the combined set of normal and abnormal training data to ensure the bottleneck learned a more comprehensive representation of human kinematics. Crucially, after establishing this broader latent space, we maintained the strict OCC protocol for the anomaly detection phase by training the EBM solely on the normal training data.

Under this modified regime, the AE performance improves to 90.1\%, and the VAE achieves 86.2\%. However, despite providing the Autoencoders with the advantage of abnormal pretraining data, both non-linear models still fall short of the linear baseline. Notably, recalculating the PCA projection using the combined normal and abnormal data yields identical performance (90.5\%), proving that the linear transformation is inherently robust and does not suffer from non-linear mode collapse. These results demonstrate that the simple linear PCA projection yields the highest performance overall, effectively preserving the structural distinction of anomalies without requiring complex Autoencoder pretraining or access to abnormal training sequences.

\begin{table}[th]
\setlength{\tabcolsep}{6pt}
\centering
\scriptsize
\caption{Comparison of linear and non-linear extractors on the UBnormal benchmark ($T=12, K=48$). \textit{Pretrain using Anomalies} indicates the projection manifold was constructed using both normal and abnormal training data, rather than strictly normal data. The linear PCA projection outperforms deep Autoencoders, even when the latter are provided with abnormal training sequences.}
\label{tab:ae_vs_pca}
\begin{tabular}{l cc cc cc}
\toprule
\textbf{Projection Method} & \multicolumn{2}{c}{\textbf{PCA}} & \multicolumn{2}{c}{\textbf{AE}} & \multicolumn{2}{c}{\textbf{VAE}} \\
\cmidrule(lr){2-3} \cmidrule(lr){4-5} \cmidrule(lr){6-7}
\textbf{Pretrain using Anomalies} & $\times$ & \checkmark & $\times$ & \checkmark & $\times$ & \checkmark \\
\midrule
\textbf{AUROC (\%)} & \textbf{90.5} & \textbf{90.5} & 89.5 & 90.1 & 84.9 & 86.2 \\
\bottomrule
\end{tabular}
\end{table}